# An Integrated System for Real-Time Student Assessment and Career Guidance Using Neural Networks in Computing Disciplines

Sakir Hossain Faruque[1] (**sakir15-3862@diu.edu.bd**), Md. Jubair Hossain[2] (**jubair.duengg@bec.edu.bd**), Sharun Akter Khushbu[1] (**sharun.cse@diu.edu.bd**)

[1]Department of Computer Science and Engineering, Daffodil International University, Dhaka-1216, Bangladesh,

[2]Department of Electrical and Electronic Engineering, Barishal Engineering College, Barishal, Bangladesh

## Abstract

Many undergraduate students in Computer Science (CS) and Software Engineering (SWE) struggle to identify suitable career paths, particularly when their academic performance, abilities, and interests do not fully align. To address this issue, this study proposes an AI-driven Student Assessment and Career Prediction System that integrates a Career Guidance Expert (CGE) system with a Web-Based Student Assessment (WBSA) platform. Within the integrated framework, CGE enhances personalized career recommendations using AI while also assisting students after graduation in identifying suitable jobs, research domains, and higher study opportunities aligned with their skills and interests. The WBSA platform further strengthens interaction between students and faculty through assessments, personalized tasks, mentorship activities, and a secure real-time chat application. The CGE system employs a Multilayer Perceptron (MLP) model trained on real-world academic and extracurricular data collected using the snowball sampling method from the students of universities, achieving a validation accuracy of 94.71% in predicting personalized career paths. A pre-survey was conducted across universities to evaluate the proposed model before deployment. The WBSA system was developed as a modern web application using technologies such as Node.js, Next.js, and PostgreSQL to ensure scalability, responsiveness, and secure data management. The overall system is supported by a secure cloud-based infrastructure, the platform provides reliable performance while assisting graduates to select suitable career path in IT sector. In addition, a post-survey involving both students and faculty was conducted to gather feedback and further improve the overall effectiveness and usability of the system.

**Keywords:** Expert System, Multilayer Perceptron, Student Assessment, Natural Language Processing, Career Guidance

# 1. Introduction

In recent years choosing a right career has become much less straightforward for undergraduates. Moreover, to the frenetic pace of globalisation, changing economies, substantial technological advancements and an increasingly competitive education system, students find it challenging to navigate their futures. In addition, the educational system is also compelled to adjust to an increasingly diverse student population with new perspectives and claims (Hirschi, 2018). Consequently, a growing number of young people feel alarmed and uncertain about their professional future (Shahiri & Husain, 2015).

This increased complexity makes it challenging for students to make sound judgments, especially if they lack expertise or access to appropriate counsel. Frequently, they depend only on parental or instructor guidance (Supriyanto et al., 2019) and pick professional options that do not match their academic achievement, confidence, personality, or particular interests, resulting in dissatisfaction and misalignment (Razak et al., 2014). Therefore, choosing a suitable career is a complex and time-consuming process (Nyamwange, 2016; Bandura et al., 2001). To address these problems, we propose a WBSA and CGE system—an AI-powered platform to assist students from the start of their university life. This system connects students to faculty mentors, assesses their skills based on interests and skill-related activities, and forecasts appropriate career paths using a neural network-based prediction model. Aside from assisting, the platform promotes a dynamic and collaborative academic atmosphere, making students' lives more disciplined, creative, and focused through meaningful mentorship and educated career preparation.

While the application of digital technologies in education is growing, there remains a significant gap in systems that integrate skills assessment with personalised career guidance, particularly for students pursuing CS and SWE. Existing solutions primarily emphasize academic grading or provide job suggestions that take no account of students' personal interests, skill sets or context of their learning. In addition, conventional counseling practices usually do not scale or provide immediate feedback, which all hampers the proper interaction of students and faculty. Therefore, the students themselves cannot track their progress and thus are not beginning to develop confidence (Dahalan & Hussain, 2010; Coleman & Penuel, 2000; Antal & Koncz, 2011). This brings about the necessity of an innovative, student-focused system that analyses a student's performance and steers them in the right career path while supporting continual growth and self-awareness.

AI is being explored to solve these challenges and many others to enhance the career advising experience. However, today, academic evaluation is separated from job prediction by most modern AI-driven systems, and students are denied the choice to correlate their real-time performances with their future paths. Most systems are still general and not tailored to a specific domain, e.g., CS and SWE. Moreover, mentorship integration is predominantly absent with most systems being based on static, rule-based approaches or one-off surveys (Supriyanto et al., 2019; Myla et al., 2019) as opposed to integrated models like multilayer perceptrons (MLP) or possessing concepts such as ability trajectories. More importantly, however, scalability and user data privacy are often ignored. Systems like Naviance, which has a student base of 7 million in 8,500 schools across 100 countries (Aadmin, 2024; Wikipedia contributors, 2025), have been popular but still depend on traditional, form-based inputs and are without real-time academic data or predictive AI technology. In a like manner, the UTS Professional Mentoring Platform (University of Technology Sydney, 2025), Qooper (Doguslu, 2024), and Year Up (Year Up United, 2025) are places where one could receive assistance with mentoring and employment; however, those platforms do not integrate the evaluation of the faculty with the AI-powered, domain-specific counsel component. The absence of these gaps indicates the enormous necessity for a single intelligent system, which not only predicts the right vocations, but also can carry on continuous academic review and implement fruitful faculty collaboration—these are the aims that this research pursues.

This work aims to create an AI-powered platform to guide CS and SWE students in their assessments and career paths. By analyzing academic performance, skills, and interests through an MLP model, it provides personalized recommendations. The platform also enables quick communication between students and faculty via WBSA, helping mentors guide students effectively. Beyond career support, it helps faculty form research groups, foster real-world projects, and promote collaboration, creating a more engaging, practical, and productive learning environment.

The main research questions that were discovered in our investigation are:

- How does using AI-driven, real-time assessments influence students' career choices and confidence?
- How does faculty involvement via a web-based platform affect student engagement, mentorship, and growth?
- How well does the system align students' skills with suitable career paths compared to traditional guidance?
- How scalable and secure is the platform for handling sensitive data and real-time assessments across institutions?
- Can it help faculty create effective research groups and project teams based on students' skills and interests?

Building on our earlier work (Faruque et al., 2025), this study introduces a unified platform for IT students that combines real-time skill assessment, personalized career guidance via an MLP model, and interactive faculty mentorship. It not only helps students make informed career choices but also supports the creation of research groups and real-world project teams, bridging academic learning and professional readiness.

## 2. Literature Review

Recent research has identified two key areas in educational support systems: assessment of students and career advising. Each contributes significantly to students' academic progress and direction in the future.

Recent advances in educational technology have resulted in various systems aimed at improving student assessment and evaluation. Wang et al. (Wang et al., 2004) developed a web-based teacher evaluation system that enhanced test administration and management using the Triple-A test analysis model. Similarly, Dahalan et al. (Dahalan & Hussain, 2010) proposed the e-ATLMS platform, which supports automated result generation and performance grading, demonstrating strong potential for online school-based assessment. Coleman et al. (Coleman & Penuel, 2000) introduced technology-based assessments to evaluate students' environmental awareness and data analysis skills, enabling the identification of both high- and low-performing learners for targeted support. Antal et al. (Antal & Koncz, 2011) presented a computerized self-assessment system that helps identify students' knowledge gaps and allows tutors to track academic progress in real classroom settings. Additionally, Guzmán et al. (Guzmán & Conejo, 2005) developed a web-based adaptive assessment system that integrates Intelligent Tutoring Systems (ITS) with SIETTE, allowing students to perform self-assessments and receive structured feedback.

Meanwhile, research has increasingly explored the use of expert systems and artificial intelligence for student career guidance. Supriyanto et al. (Supriyanto et al., 2019) examined how expert systems can support educational, evaluative, academic, and employment decisions, showing that they enhance learning outcomes, specialization, performance, and self-assessment. Huang et al. (Huang, 2022) proposed an AI-driven career counseling system for university students, which improved recommendation performance and employability-focused job matching, as reflected in higher AUC values. Mehraj et al. (Mehraj & Baba, 2019) highlighted the growing role of automation and digitalization in career guidance, demonstrating the effectiveness of techniques such as machine learning, genetic algorithms, fuzzy logic, and neural networks, and suggested integrating AI with civic education to guide students' employment journeys. Crişan et al. (Crişan et al., 2015) found that students' skills and knowledge often do not align with their expected jobs, identifying common barriers in career decision-making and emphasizing the need for effective counseling. Myla et al. (Myla et al., 2019) developed an expert system that helped students select suitable career paths based on questionnaires, outperforming traditional counseling by considering academic performance and recommending appropriate study branches.

## 3. Development of an expert system

It is challenging for students who are transitioning from high school to college, a critical period that shapes their future professional paths. Considering abilities, skills, personal and social responsibilities, making the right choice is essential, and a career guidance expert system can play a key role in supporting students through this decision-making process (Kazi & Akhlaq, 2017).

### 3.1. Expert System

An expert system (ES) is a computerised interactive tool that relies on factual data and expert knowledge (heuristics) to solve complex decision-making tasks. It simulates a human expert's reasoning process by using a set of rules, usually in the form of IF-THEN statements, to analyse information, ask pertinent questions, and give explanations or alternative answers, making it a dependable tool in disciplines such as health, education, and agriculture (Mehraj & Baba, 2019; Kazi & Akhlaq, 2017). In Figure 1, we have shown how humans interact with an expert system.

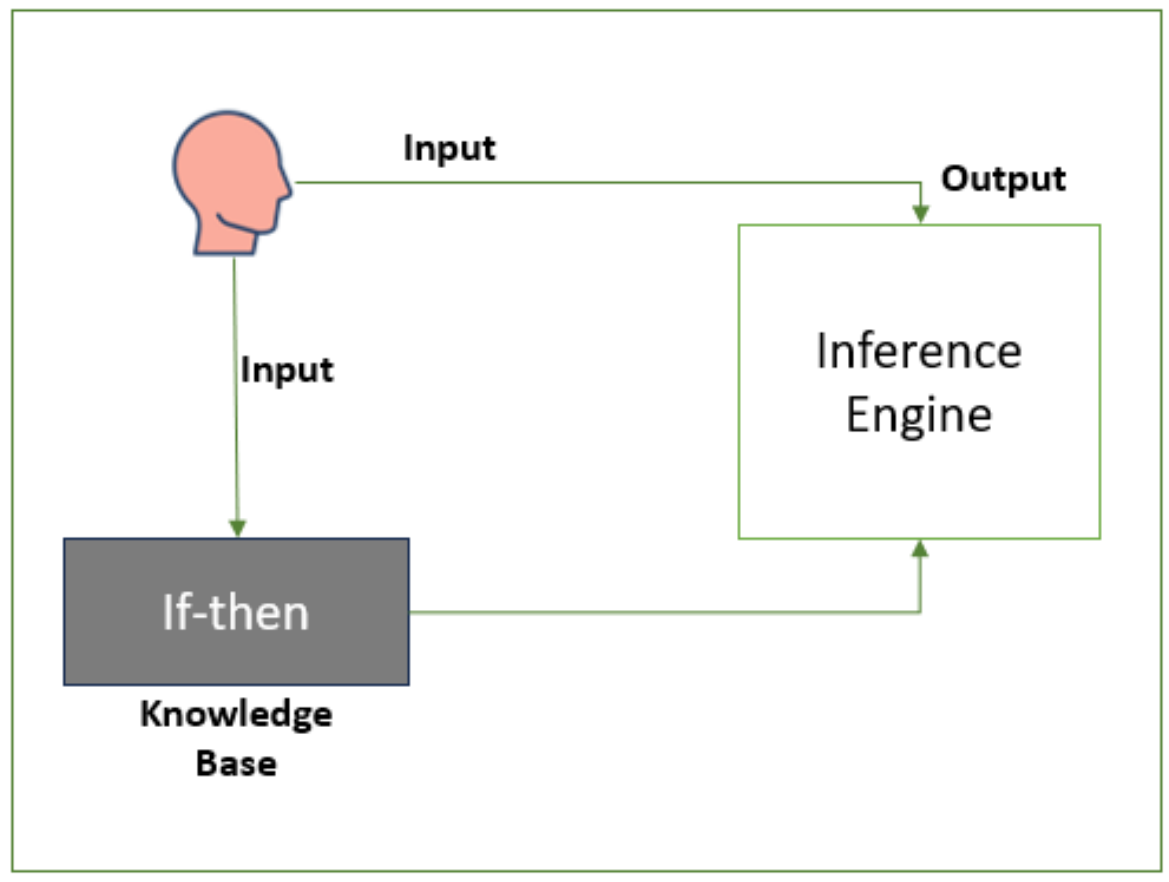

Figure 1: Human Interaction with Expert System

#### 3.1.1. Career guidance expert system (CGE):

In this article, we focus on decision-making expert systems, particularly those designed for career guidance. We have published previous research on career guidance and related technologies where we demonstrated the importance of career advice in students' lives, especially for IT (CS and SWE) students (Faruque et al., 2025). The career guidance expert system is an AI-based recommendation system that includes key components such as a user interface, expert system database, knowledge acquisition tools, and an inference mechanism (Faruque et al., 2025; Myla et al., 2019). Many students often choose careers without proper advice, which may not align with their interests, skills, or future satisfaction (Reddy, 2021). Expert systems, a branch of artificial intelligence, are increasingly being utilised to address this issue by measuring students' academic achievement, extracurricular activities, personality traits, and preferences (Reddy, 2021; Rangnekar et al., 2018). This study emphasises the importance of creating a web-based expert system to deliver precise, personalised career guidance. As students today no longer rely solely on parents or teachers for career decisions, a competent, credible, and independent expert system is essential. Such systems are already being widely used in various domains, including academic counseling, vocational guidance, and educational evaluation (Supriyanto et al., 2019).

### 3.2. Development of career guidance expert system

Creating a career advice expert system requires many critical phases, beginning with gathering and processing student data and progressing to training AI models, analysing findings, choosing the best-performing model, and testing it in real-world circumstances to ensure accurate and valuable suggestions (Faruque et al., 2025).

**Data collection and analysis (Faruque et al., 2025):** We collected data from CS and SWE students at top universities in Bangladesh using snowball sampling and Google Forms. We focused on their academic backgrounds and extracurricular activities to understand career preferences. Analysis revealed that only 16.4% of students aimed for software development roles. Figure 2 shows the variety of career paths chosen by the students, highlighting the diversity of their aspirations.

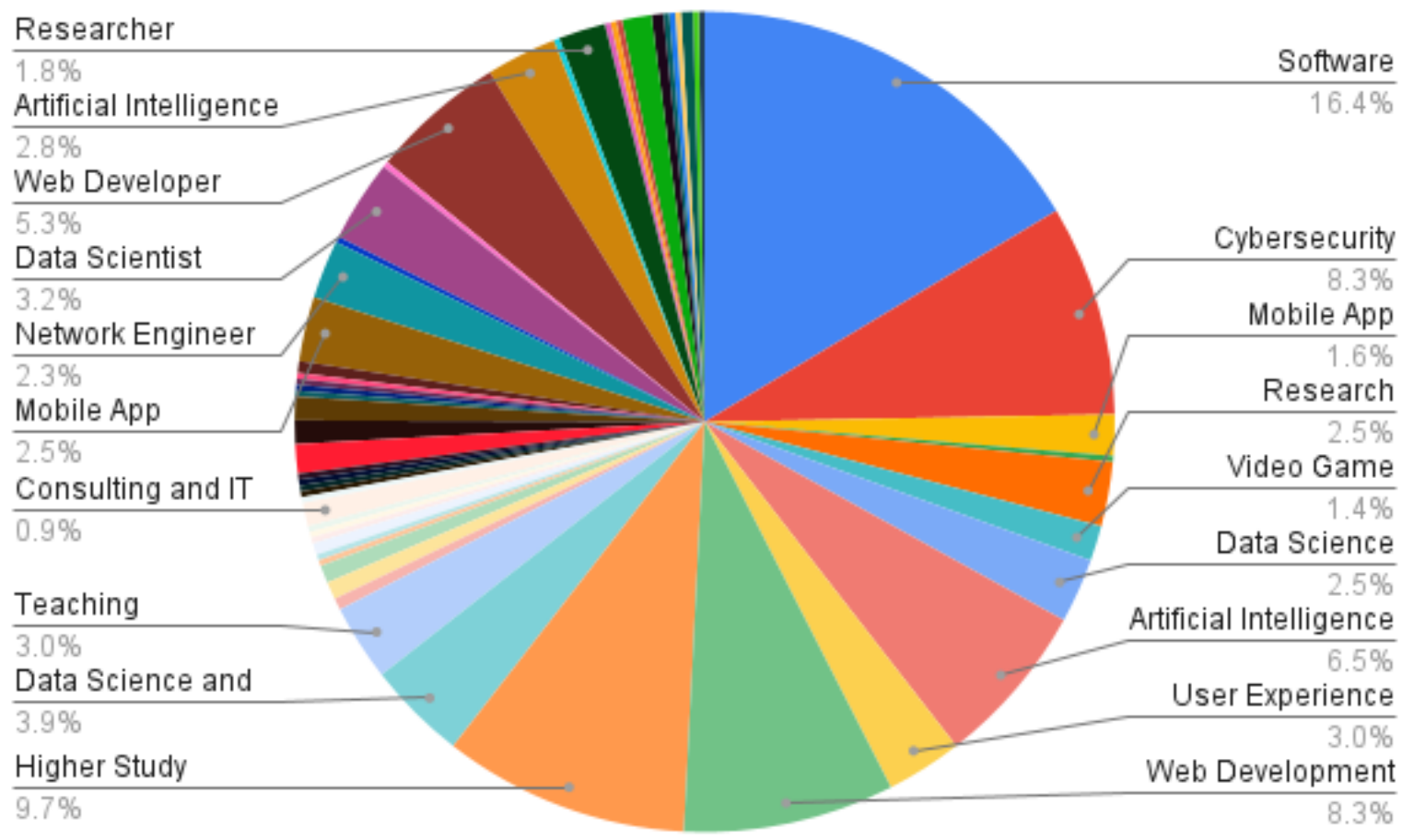

Figure 2: Statistical distribution of students' preferred career fields in percentage terms, based on (Faruque et al., 2025)

**Data processing (Faruque et al., 2025):** We preprocessed the raw data for machine learning by cleaning it, selecting relevant features, removing unnecessary columns, converting text to lowercase, eliminating stop words, and applying label encoding and text vectorization. The dataset was then split into training and testing sets with shuffling to ensure unbiased performance, and Table 1 shows the labels assigned to each career class for supervised training.

Table 1: Encoded representation of career field classes applied in the proposed model, based on (Faruque et al., 2025)

| Label | Master Field |
|---|---|
| 0 | AI |
| 1 | DS |
| 2 | DEV |
| 3 | SEC |
| 4 | SDE |
| 5 | UI / UX |

**Apply machine learning (Faruque et al., 2025):** To build an accurate career guidance expert system, we applied both traditional supervised learning and deep learning (DL) techniques. We used DT, SVM, LR, KNN, and NB for the classical ML models. We trained models like CNN, MLP, and LSTM on the DL side, incorporating multiple dropout and dense layers to improve learning performance and reduce overfitting.

**Result analysis and model selection (Faruque et al., 2025):** After training, we evaluated the models using metrics like accuracy, precision, recall, F-measure, AUC-ROC, and learning curves. The MLP outperformed all others with 94.71% validation accuracy and low generalization error, making it our choice for its strong performance, efficient multiclass classification, and scalability on large datasets.

**Model Input Output (Faruque et al., 2025):** In Table 2, we present the MLP model's career recommendations for CS and SWE students based on their input data. The model outputs percentages indicating the likelihood of success in various career fields, and we focus on the top 2% of these predictions. To make the results more meaningful and student-friendly, Table 3 shows how we refined the output, labeling the highest percentage as the "primary goal" and the second highest as the "secondary goal," helping students clearly understand their most suitable career paths.

Table 2: Example of human input features and machine-predicted career categories generated by the MLP model modified from (Faruque et al., 2025)

| SL | Human Input Skills | Predicted Career Categories with Probability (%) |
|---|---|---|
| 1 | AI, ML, DS, WD, SDE, GD | SDE (56.94%), AI (20.05%), DS (17.00%), UI/UX (2.68%), DEV (1.70%), SEC (1.63%) |
| 2 | ML, DS, CS, ISA | SEC (52.13%), SDE (21.38%), AI (13.99%), DS (10.78%), UI/UX (1.41%), DEV (0.32%) |
| 3 | AI, ML, DS, DSA, TPA | AI (49.45%), DS (40.58%), SDE (7.98%), SEC (0.99%), UI/UX (0.87%), DEV (0.15%) |

Table 3: Career pathway mapping based on machine-predicted categories modified from (Faruque et al., 2025)

| SL | Machine Predicted Output | Example Industrial Roles | Potential Research Fields | Recommended Master's Programs |
|---|---|---|---|---|
| 1 | **SDE** (Software Development) | Software Developer, Project Manager, System Analyst, QA Engineer, Product Manager, IT Consultant, Startup Founder | Software Engineering, User Research, Research Software Engineering | Computer Science, Software Engineering, Information Systems, Cybersecurity, Data Science, Human–Computer Interaction, MBA, Project Management |
| 2 | **AI** (Artificial Intelligence) | AI Engineer, Machine Learning Engineer, Data Scientist, Robotics Engineer, NLP Engineer, AI Consultant, AI Architect, AI Product Manager | Natural Language Processing, Computer Vision, Reinforcement Learning, Generative AI, Explainable AI, AI Ethics, Robotics | Artificial Intelligence, Computer Science with AI, Data Science with AI, NLP, Computer Vision, Cognitive Science, AI in Business |
| 3 | **DS** (Data Science) | Data Scientist, Data Engineer, Data Analyst, BI Analyst, Data Architect, Data Consultant, ML Engineer | Big Data Analytics, Deep Learning, Predictive Analytics, Time Series Analysis, Graph Data Science, Data Privacy, AutoML | Data Science, Statistics with Data Science, Business Analytics, Machine Learning, Data Engineering |
| 4 | **UI/UX** (User Interface and Experience) | UX Designer, UI Designer, Product Designer, Interaction Designer, UX Researcher, Usability Analyst, Design Systems Manager | UX Research, Cognitive Psychology, Ethnographic Research, Human Behaviour Studies | Human–Computer Interaction, Interaction Design, UX Design, Industrial Design, Graphic Design, Visual Communication |
| 5 | **DEV** (Software Development – General) | Full-Stack Developer, Backend Developer, Frontend Developer, Mobile App Developer, DevOps Engineer, Blockchain Developer | Web Development, Mobile Application Development, Game Development, Blockchain Research | Software Engineering, Web Development, Mobile App Development, Game Development, Blockchain Technology |
| 6 | **SEC** (Cybersecurity) | Security Analyst, Ethical Hacker, Security Architect, | Cryptography, Network Security, Cloud Security, | Cybersecurity, Information Security, Network Security, |

| SL | Machine Predicted Output | Example Industrial Roles | Potential Research Fields | Recommended Master's Programs |
|---|---|---|---|---|
| | | Cybersecurity Consultant, Incident Response Analyst, Forensic Analyst | IoT Security, Digital Forensics, AI Security | Digital Forensics, Ethical Hacking, Cloud Security |

**Real-world testing (Faruque et al., 2025):** We developed a user-friendly interface to help students interact with the CP expert system and provide comments on its performance. This hands-on experience enabled us to analyse how effectively the technology performs in real-world scenarios.

# 4. Web-Based Student Assessment (WBSA)

Web-based student assessment connects teachers and students on a shared online platform, making it easier to create and deliver tests (Wang et al., 2004; Dahalan & Hussain, 2010). By providing real-time feedback, it helps students understand their strengths and weaknesses while fostering a more engaging and interactive learning experience (Coleman & Penuel, 2000).

## 4.1. Architecture of web-based student assessment

This section describes the development of the WBSA system, which is primarily intended to involve students in the IT sector, namely those studying CS and SWE. The system also actively incorporates faculty members from CS and SWE departments at several Bangladeshi institutions, creating a collaborative atmosphere where students may interact with mentors who share their academic and professional interests.

**Student profile:** A student starts by registering in the system and providing the required information, as detailed in Table 4. Then they receive a verification email with a unique validation link. Once verified, they can access their personal profile (Figure 3) to showcase their skills, interests, and activities, and enhance it with relevant links like GitHub or GitLab to demonstrate their expertise (Figure 4).

Table 4: Student registration form data

| SL. | Field Name | Expected Value |
|---|---|---|
| 01 | First Name, Middle Name and Last Name | Student's full name |
| 02 | Student ID | The unique student ID assigned by the university |
| 03 | Gender | Gender is either Male or Female |
| 04 | University Name | The full name of the current university/institution |
| 05 | Student Email | The official university-issued student email address |
| 06 | Password and Confirm Password | A strong password (minimum six characters) and confirm it |
| 07 | Upload | A recent formal passport-size photograph in image format (PNG, JPG) |

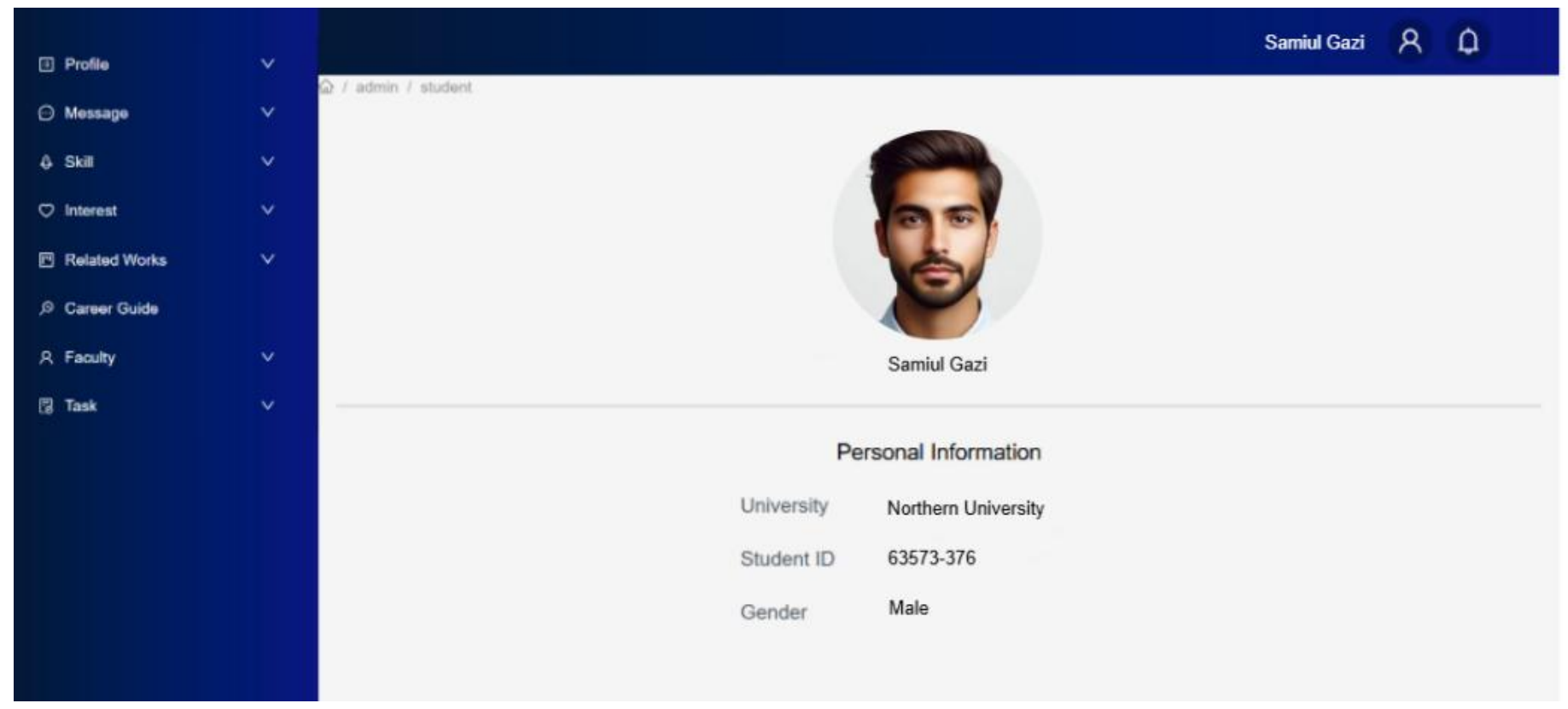

(a) Student profile data

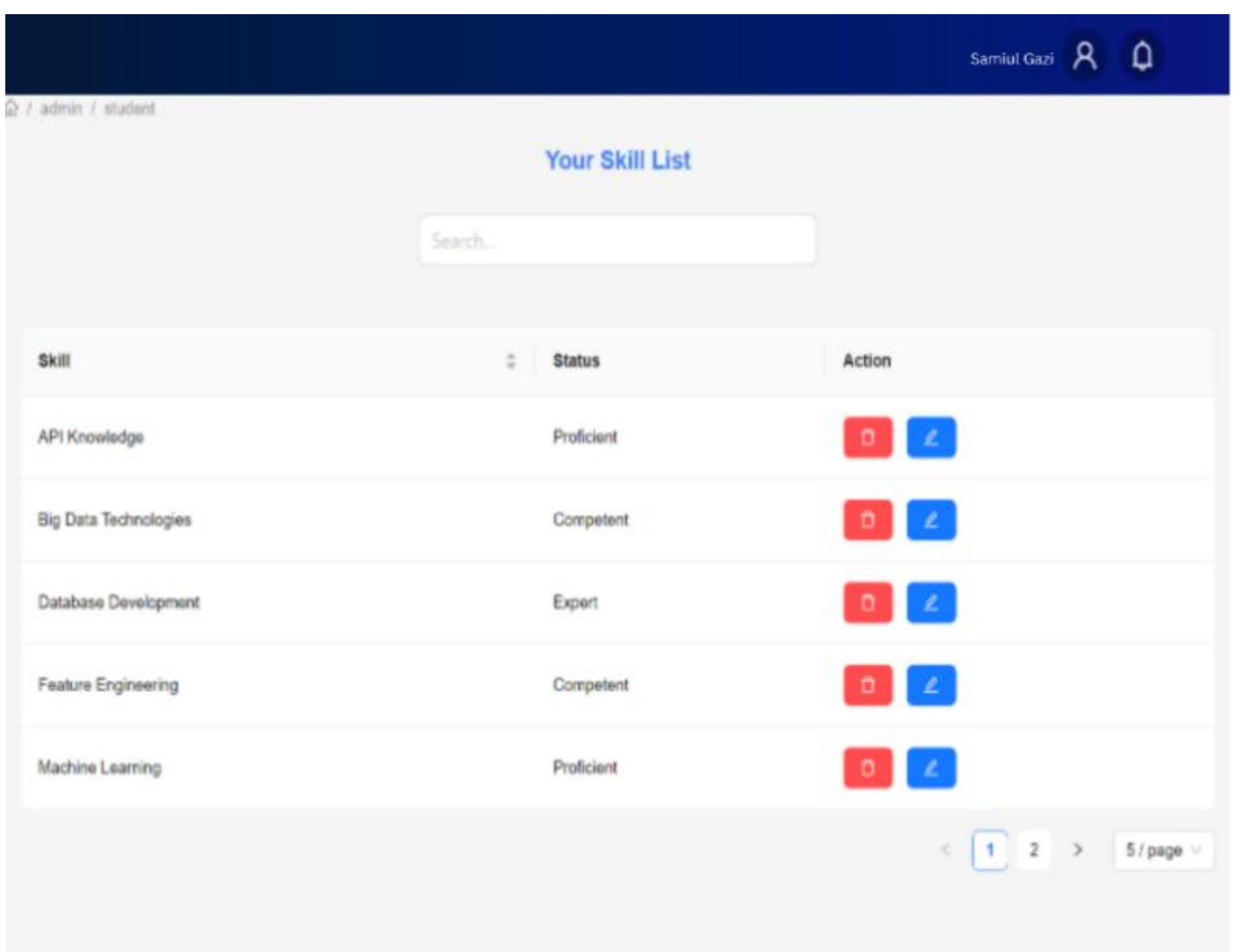

(b) Student skill fields

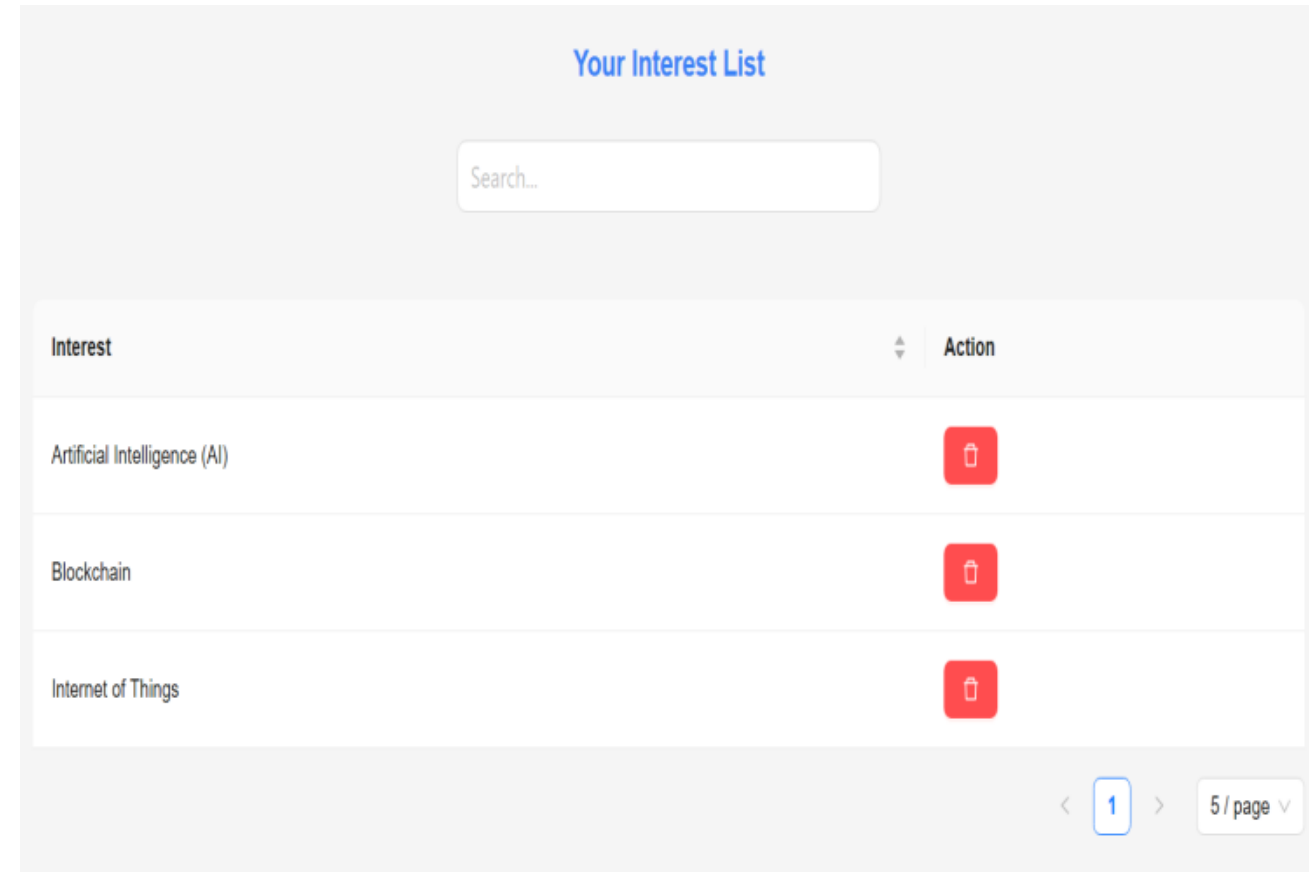

(c) Student interest fields

Figure 3: Student profile data with skill and interest

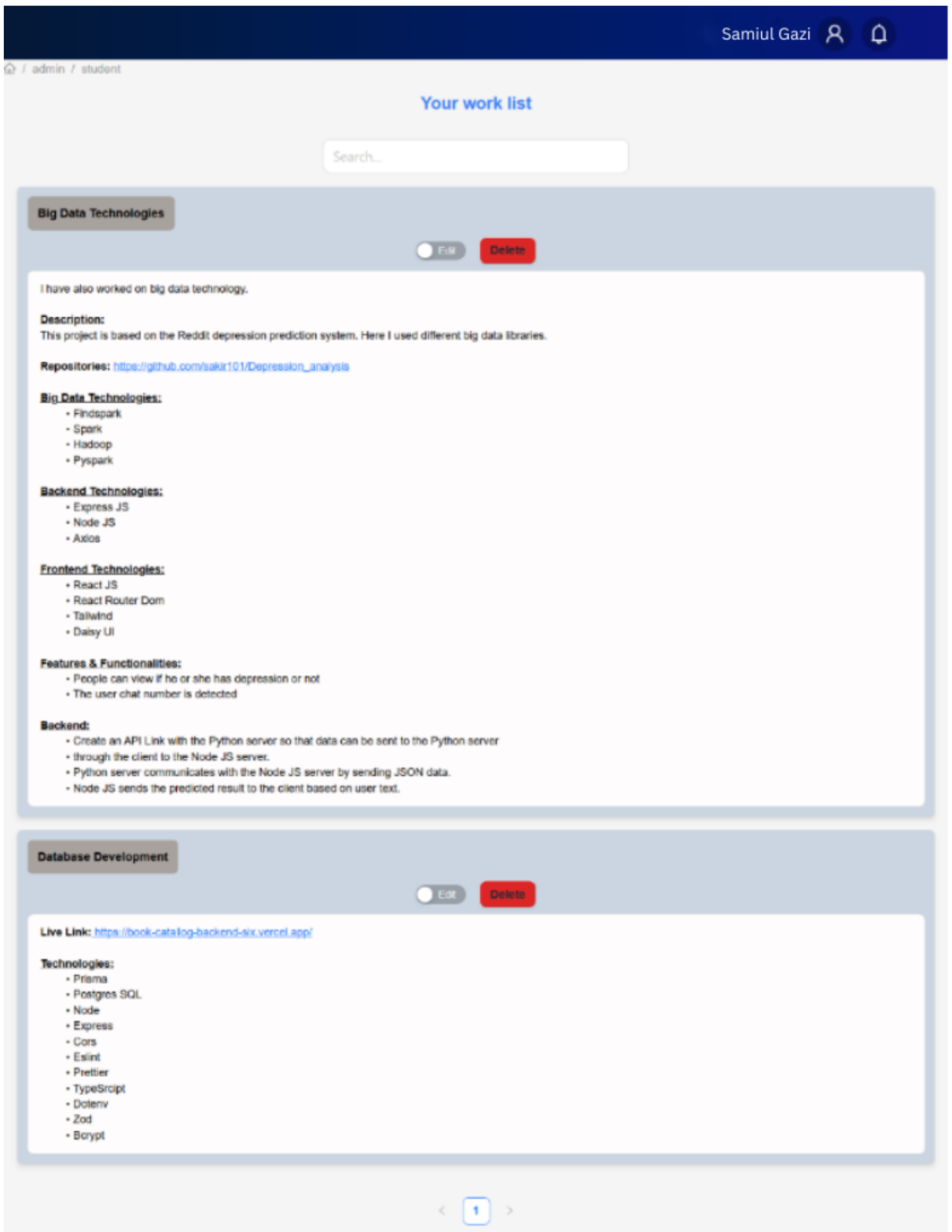

Figure 4: Student skill-based relevant work fields

The personalized profile helps the system recommend faculty members aligned with the student's academic and career goals, facilitating connections with suitable mentors (Figure 5). Once connected, students receive assignments, submit work, get feedback, and communicate with mentors in real time, with Figure 6 showing their task pages for seamless interaction.

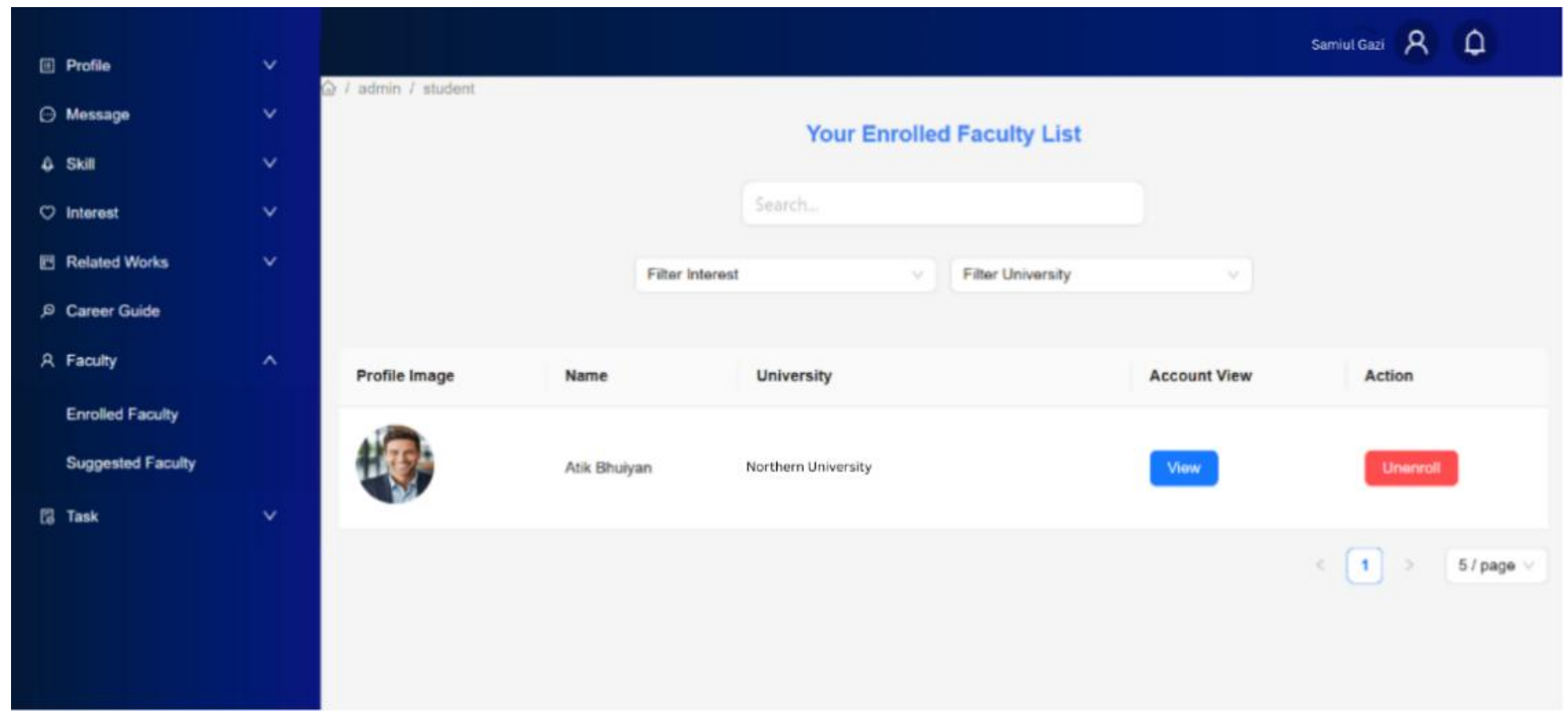

Figure 5: Student enrolled under faculty page

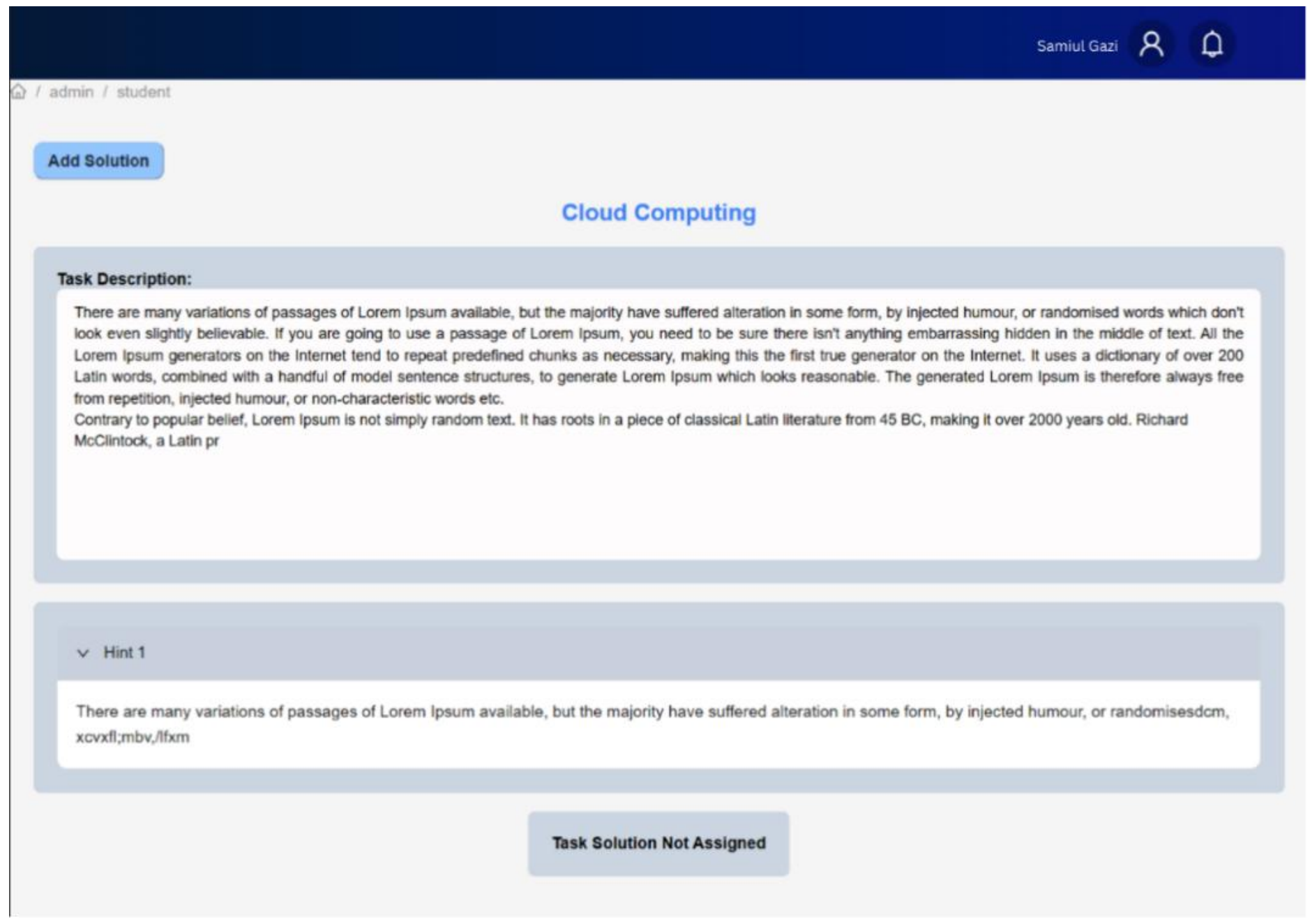

Figure 6: Student assigned task page view

**Faculty Profile:** Faculty members participate actively in the system by first enrolling using a thorough form (shown in Table 5), where they submit personal and academic information. Following registration, the administrator verifies their information, and they are sent an email with a confirmation link for account validation. Once confirmed, faculty have access to their profile (Figure 7), which they may customise by include their areas of expertise and relevant research work (Figure 8).

Table 5: Faculty registration data

| SL. | Field Name | Expected Value |
|---|---|---|
| 01 | First Name, Middle Name and Last Name | Faculty's full name |
| 02 | Employee ID | The unique employee ID assigned by the university |
| 03 | Gender | Gender as either Male or Female |
| 04 | University Name | The full name of the current university/institution |
| 05 | Faculty Email | Can provide any mail (personal or official) |
| 06 | Password and Confirm Password | A strong password (minimum 6 characters) and confirm it |
| 07 | Upload | A recent formal passport-size photograph in image format (PNG, JPG) |

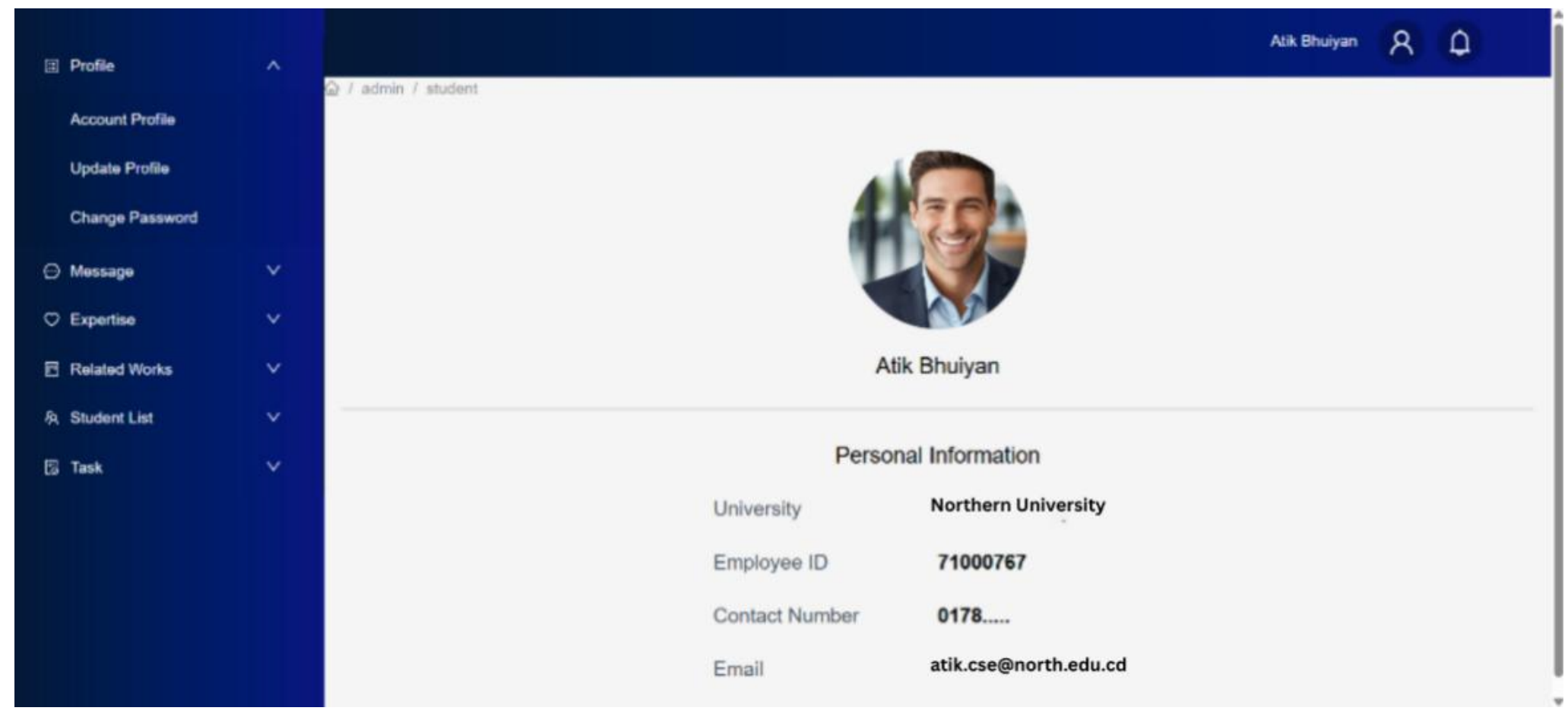

Figure 7: Faculty profile page

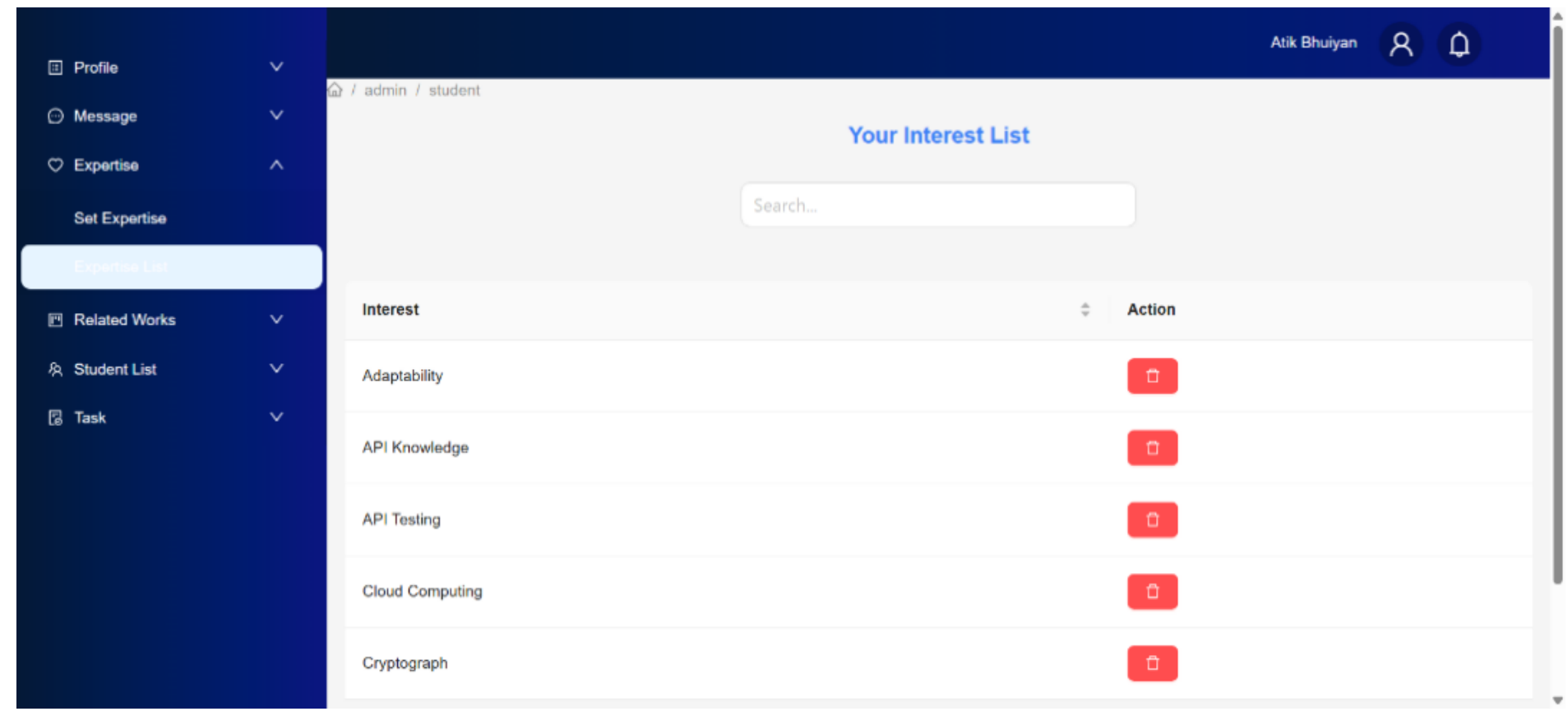

(a) Faculty expertise fields

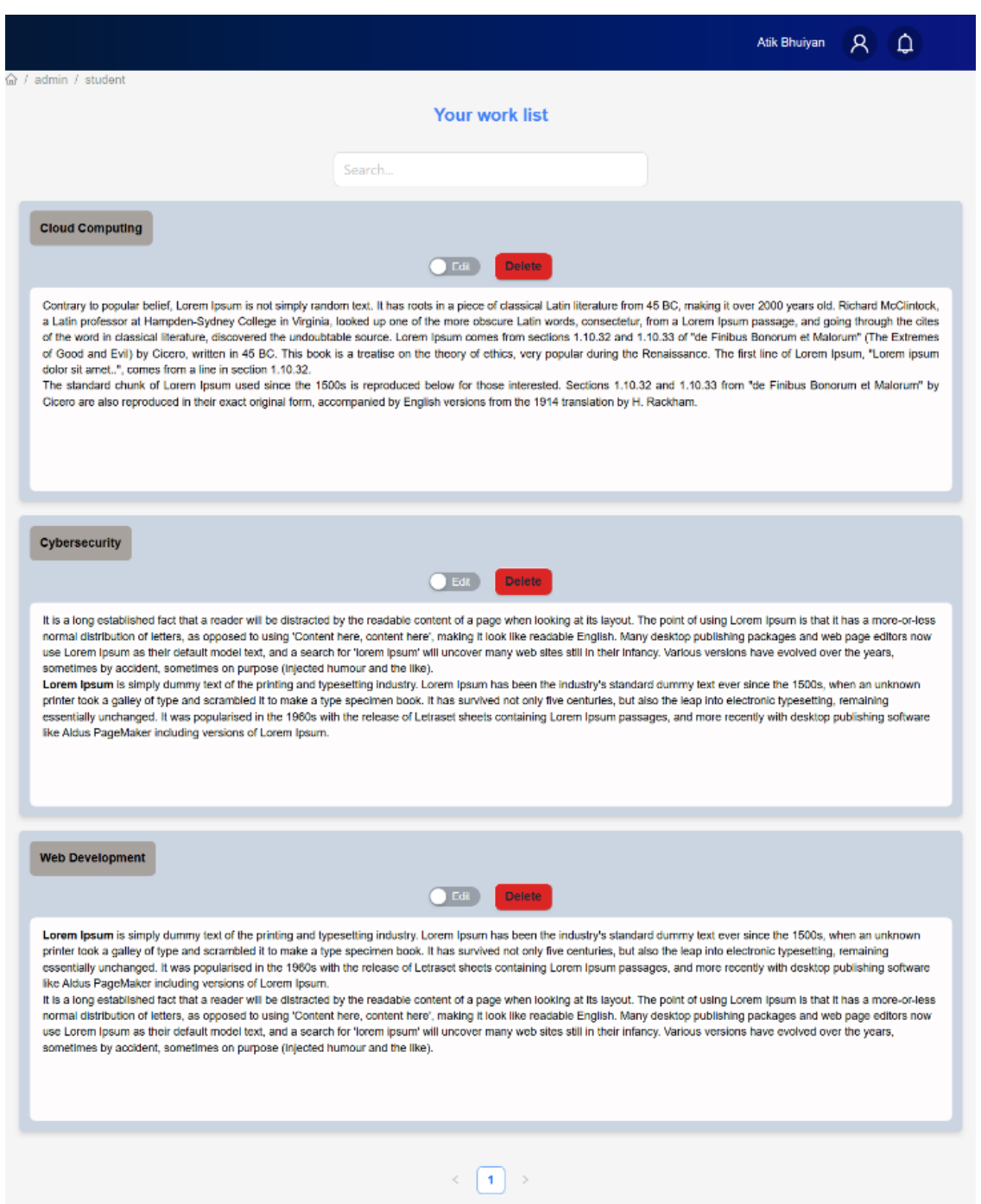

(b) Faculty expertise-based relevant work fields

Figure 8: Faculty expertise and work fields pages

The tool gives faculty a clear overview of all students in their mentorship program and allows real-time one-on-one conversations to boost engagement (Figure 9). Instructors can also create and assign academic activities, review submissions, and foster a collaborative learning environment, providing a personalized and dynamic mentoring experience for both students and faculty (Figure 10).

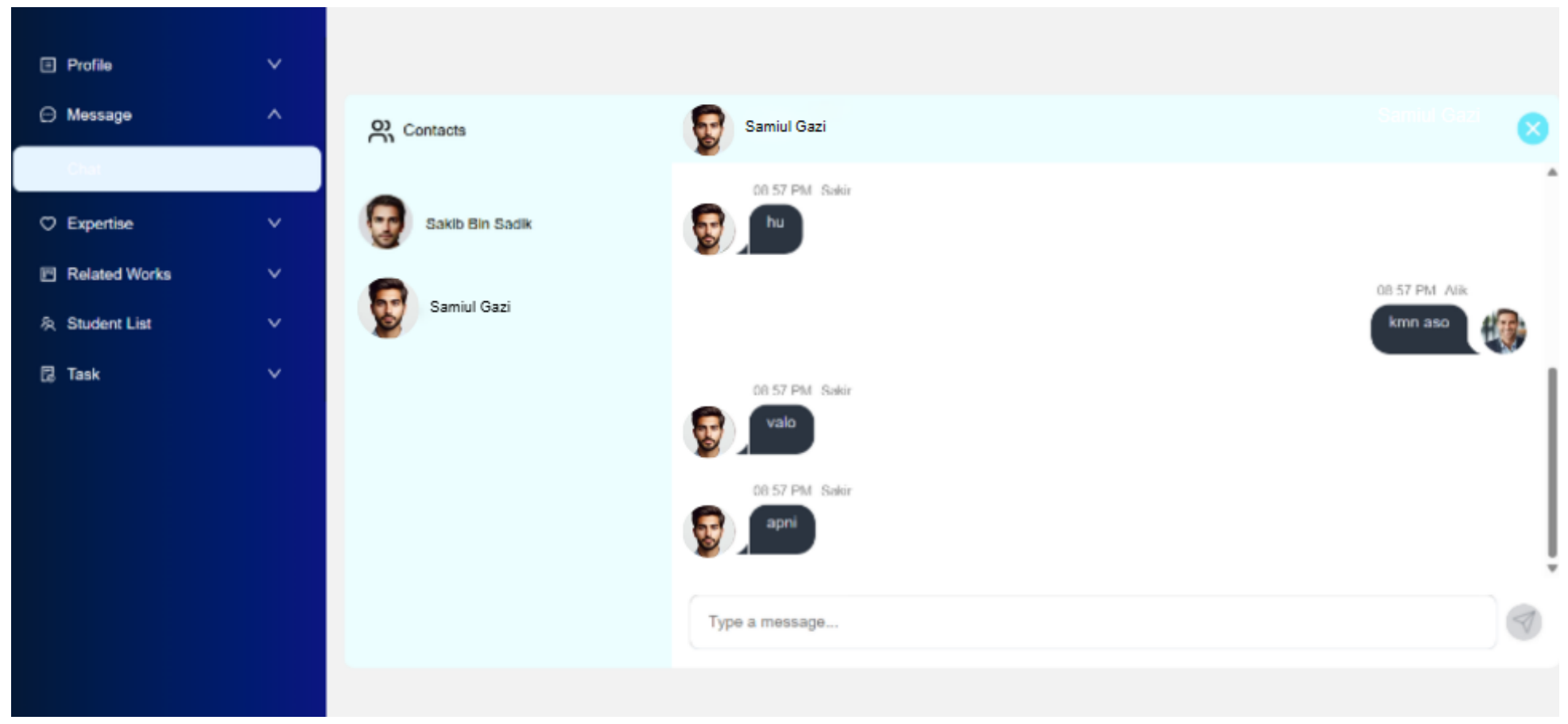

Figure 9: Faculty making real time chat with student

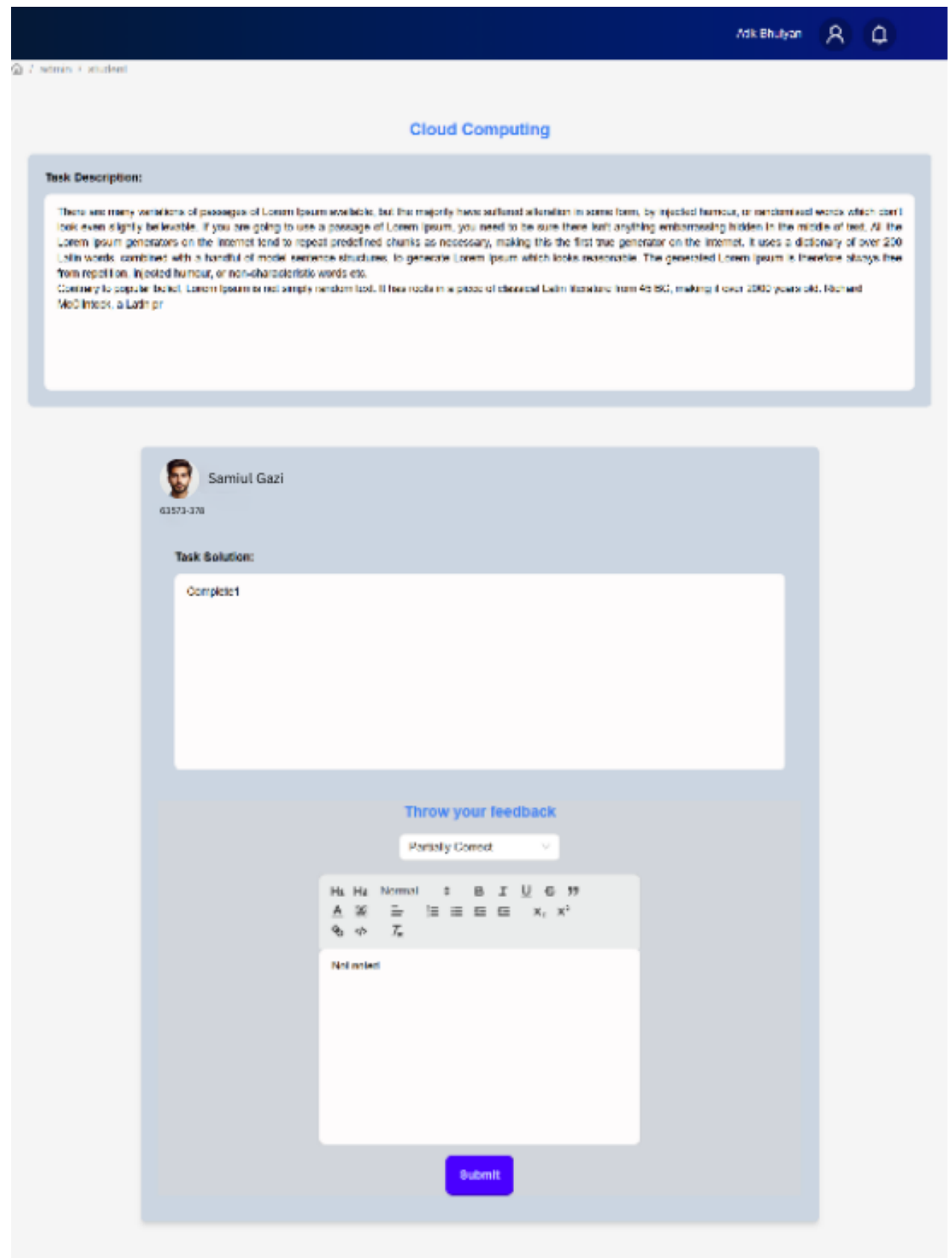

(a) Evaluate task page

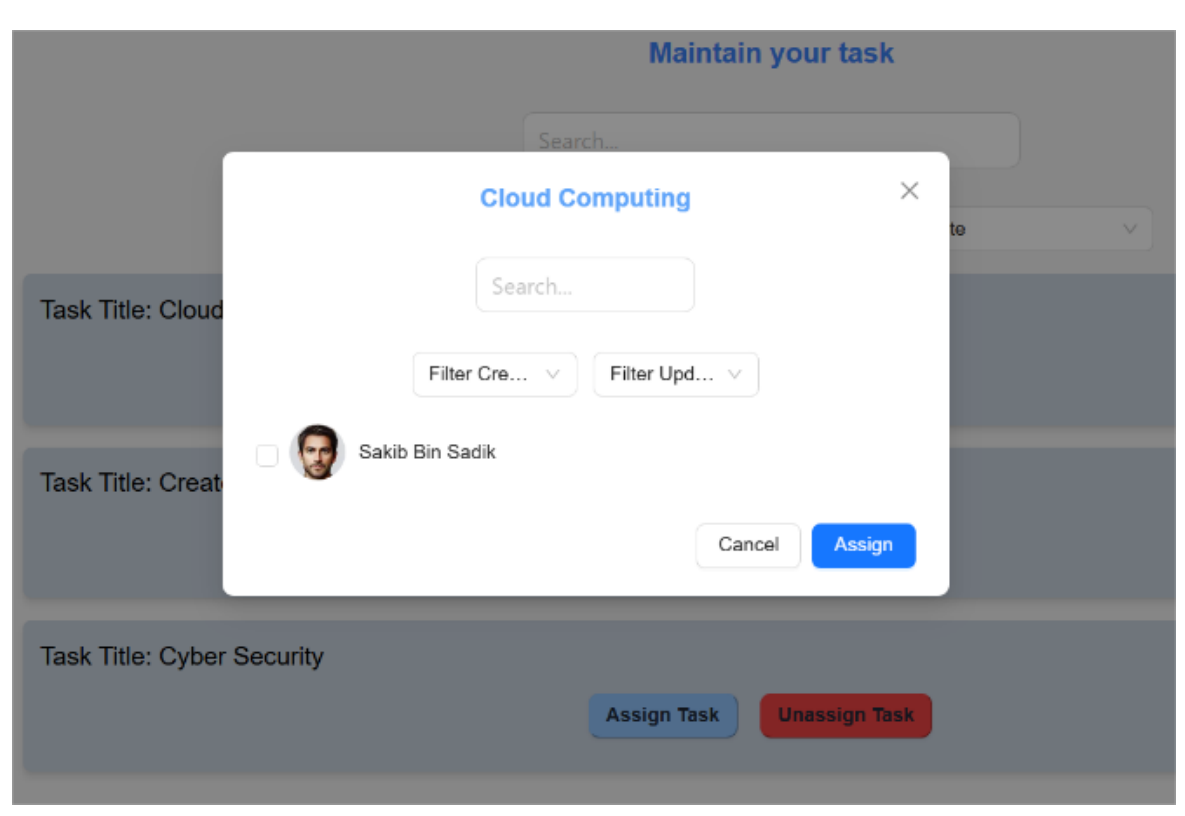

(b) Assign task to student

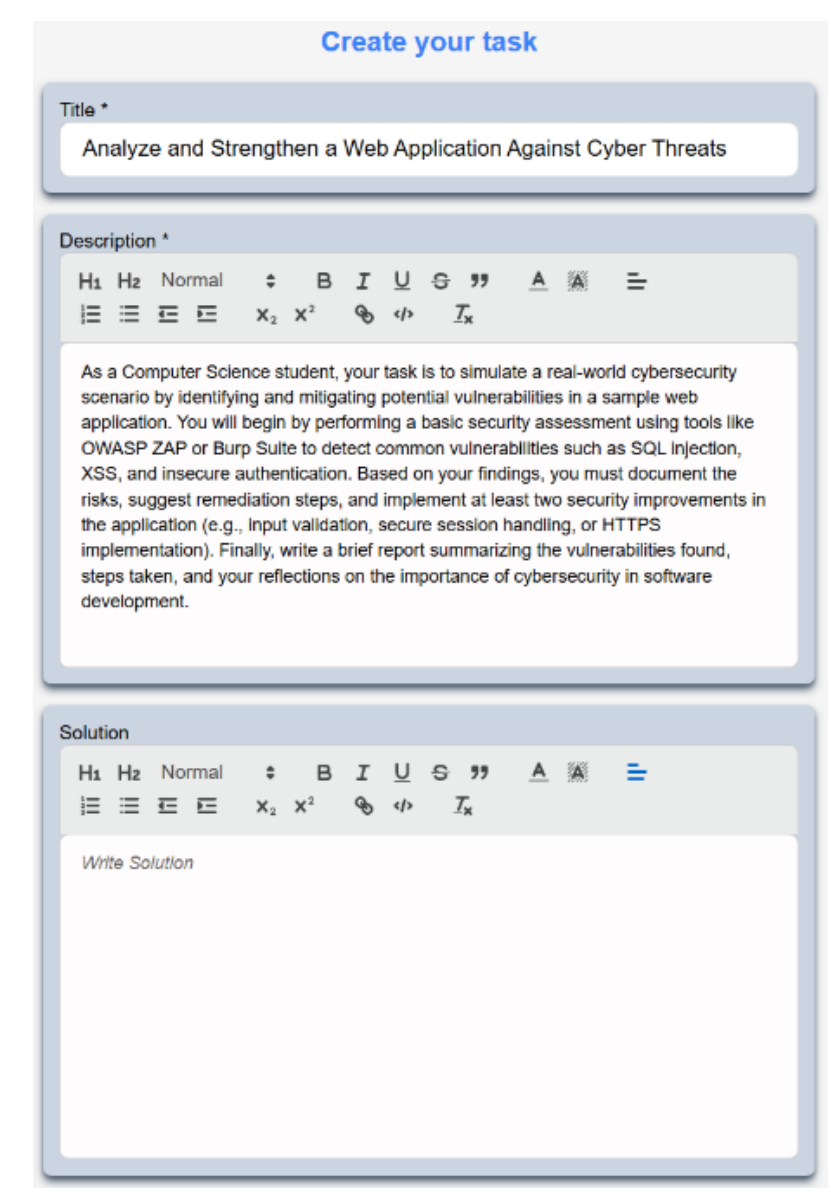

(c) Create task for student

Figure 10: Task evaluation, task assign and create new task to student faculty pages

**System Administrator:** The System Administrator manages the platform's functionality, overseeing user accounts, job listings, courses, and training centers while ensuring security and smooth backend operations (Figure 11). They also monitor for errors, maintain data accuracy, and provide user support, making their role essential for the platform's reliability and user experience.

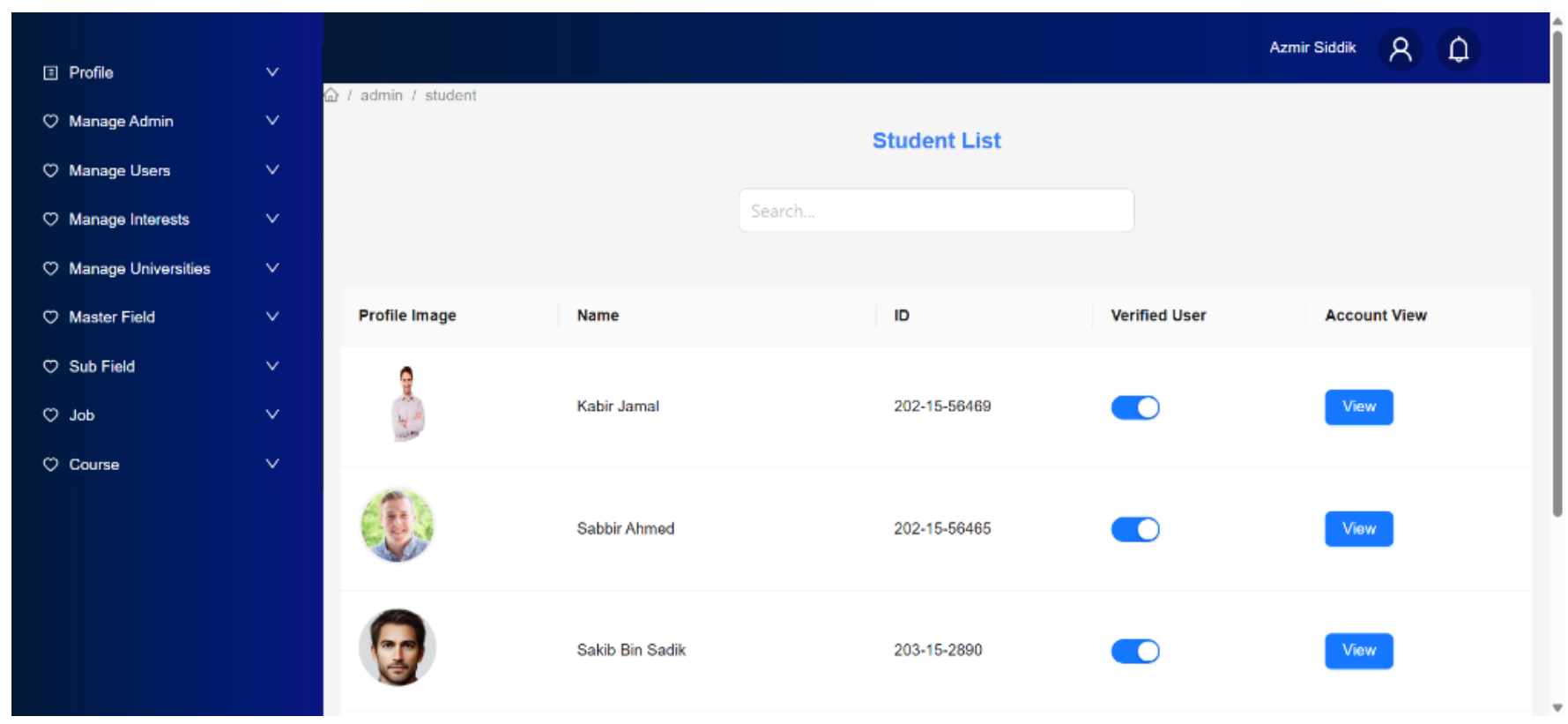

(a) Manage user page

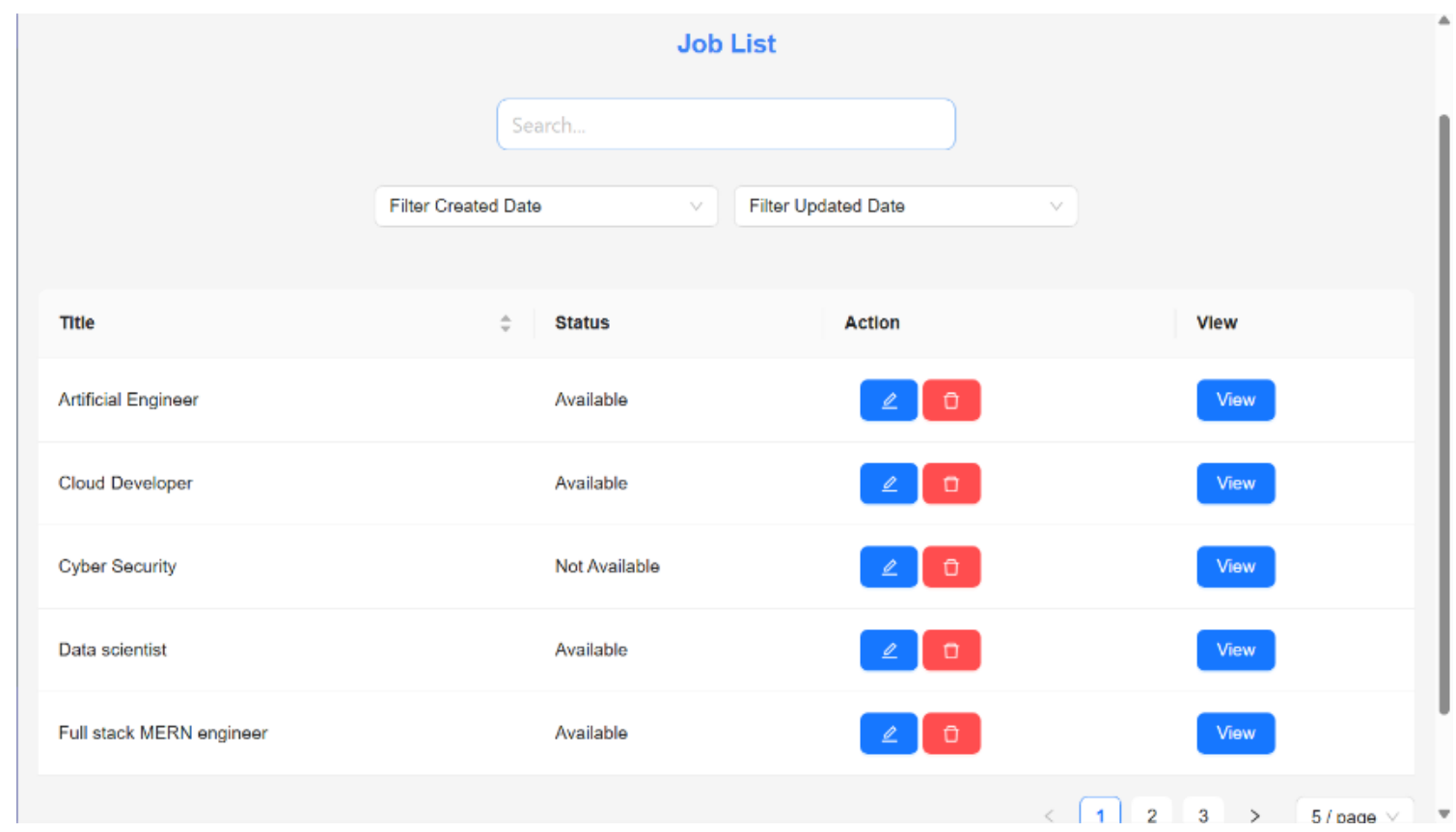

(b) Mange job posting page

Figure 11: Admin dashboard

## 4.2. Features and functionalities

We have discussed several features and functionalities of the system which makes the system more efficient to use.

### 4.2.1. Student-Faculty matching algorithm

The system applies an intelligent algorithm to assist students in choosing the most appropriate and approachable faculty members. It assesses the student's skills, interests and areas of work and compares them to academic expertise and research topics. Priority is given to disciplines where the student demonstrates outstanding proficiency and aligns with the faculty's specialisation. Furthermore, faculty members at the student's institution are scored higher. The application then provides a prioritised list of suggested instructors on the student's dashboard, making it more straightforward for them to connect with the appropriate mentors. Figure 12 depicts a schematic of the complete procedure for finding the best suitable faculty.

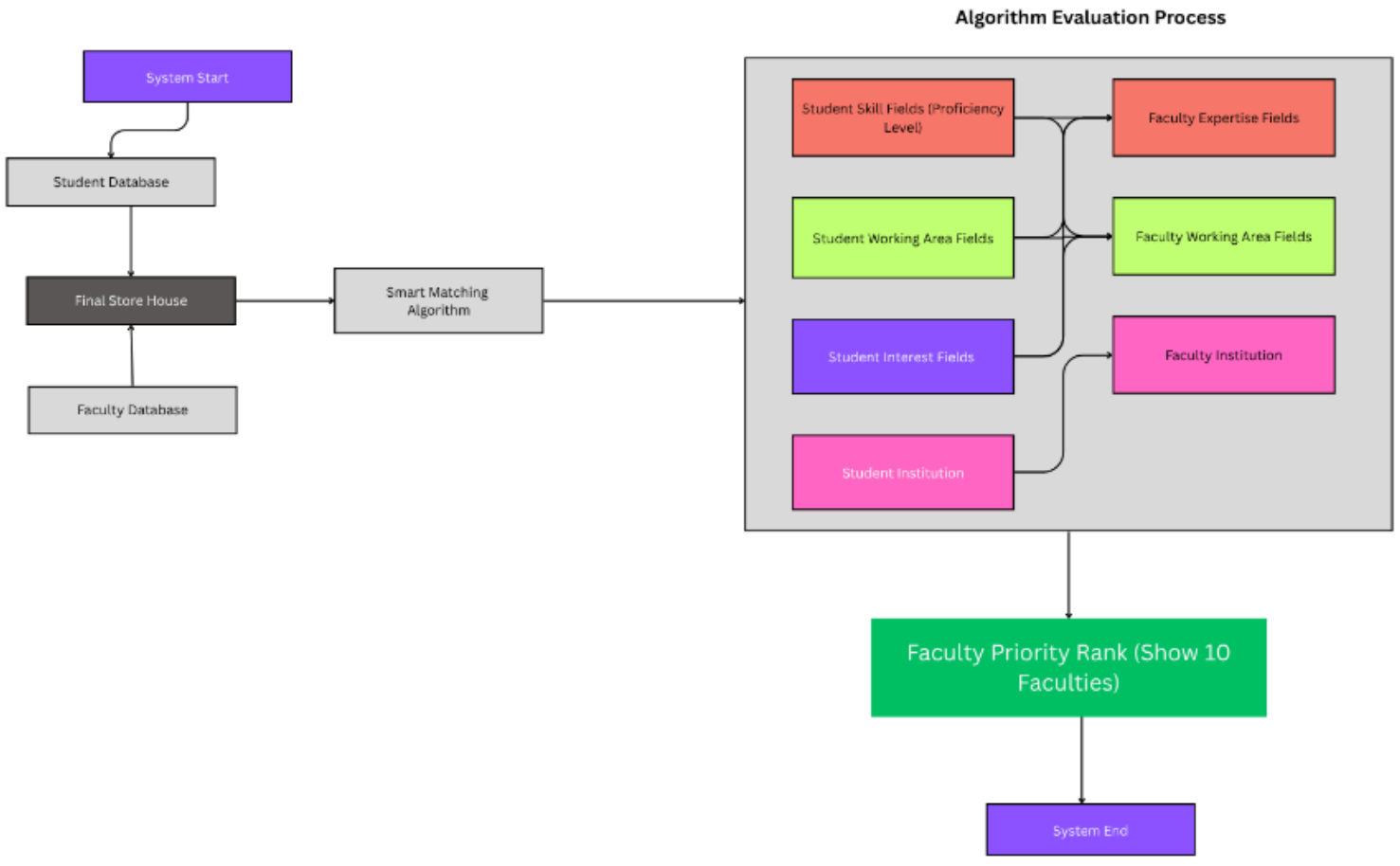

Figure 12: Optimized faculty match flowchart

#### 4.2.2. Real time search and filtration

We added real-time search capabilities to all view items' features to help users identify particular information inside enormous databases. This helps users to rapidly find entries as they type, without having to navigate through long lists. Furthermore, additional filtering tools allow users to limit down results based on common features or properties. Together, those features improve the user experience, increase efficiency, and promote the system's scalability by assuring quick and accurate data retrieval.

#### 4.2.3. System security

We take security seriously. The system uses several tough layers of protection to keep everyone safe. Students sign up with their real university email, and we double-check it with a confirmation link. Faculty accounts get even more scrutiny—admins look them over and approve them by hand. To get into the dashboard or use any features, user needs a special token, and we encrypt every password in the database. Even if someone breaks in, your account details stay protected. The admin panel keeps an eye on what users do and say, too, so trust stays high. Figure 13 shows how these three layers of security work together to protect everyone.

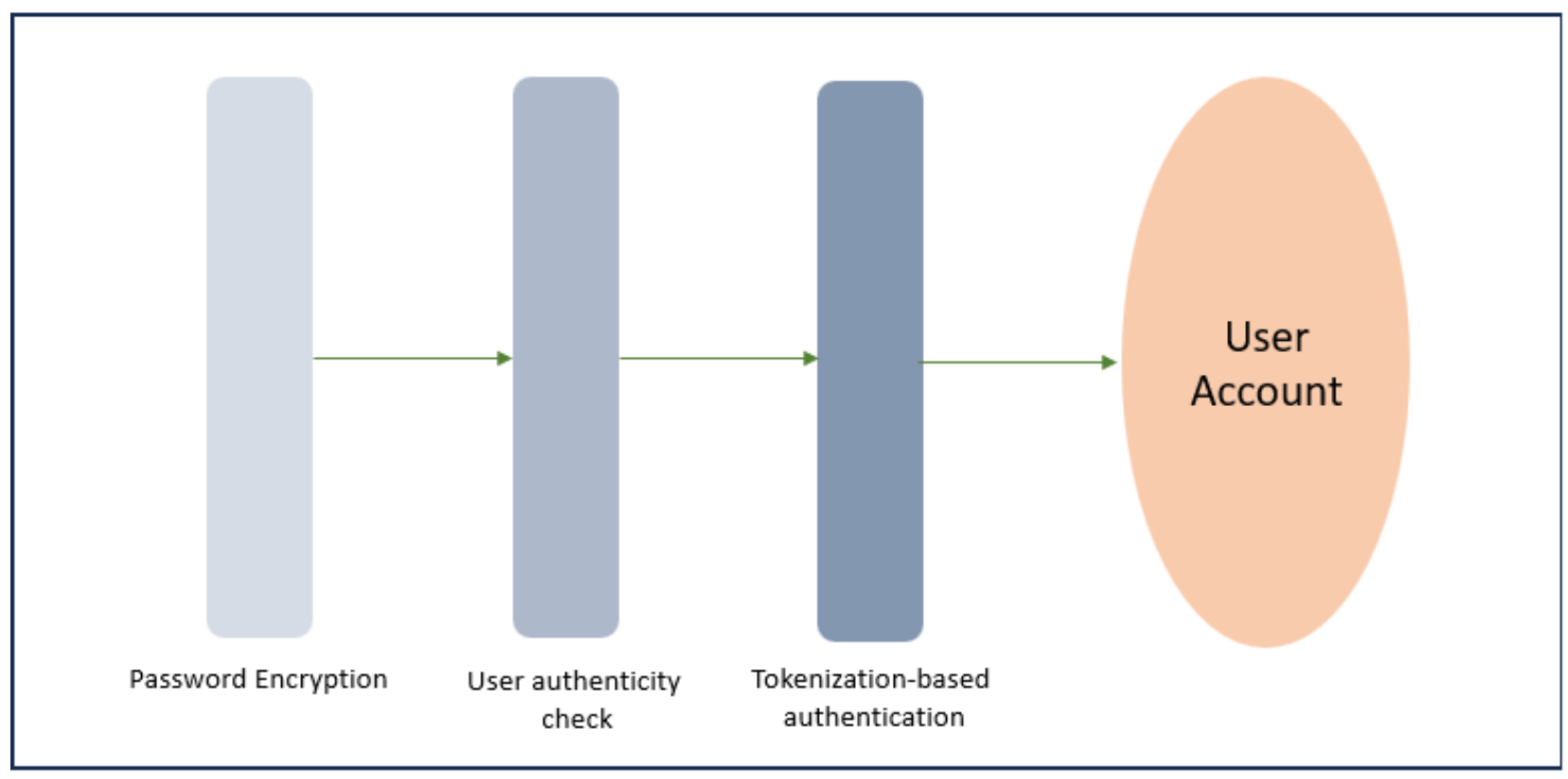

Figure 13: Three layers of security for each user account

## 5. System Integration

The core feature of our system is the seamless integration of the CGE system with the WBSA platform, offering CS and SWE students an AI-driven, user-friendly tool to explore the best career paths early in their studies. Unlike previous research (Antal & Koncz, 2011; Supriyanto et al., 2019; Huang, 2022), our approach provides a practical, comprehensive solution that guides students based on their skills and interests while helping them understand their weaknesses to make more confident career choices.

## 5.1. Integration procedure

To generate accurate career predictions, we selected the best-performing model on unseen data, with the MLP neural network outperforming all others and becoming our primary choice. This model was deployed on a Python server using Flask to enable smooth communication between the CGE system and the WBSA server. When a user requests career guidance, the query is first sent to the WBSA server, forwarded to the CGE server for processing, and the prediction is returned through the same path to be displayed in the user interface. Figure 14 illustrates the data flow between the user and both servers, while Figure 15 shows the career prediction interface from the user's perspective.

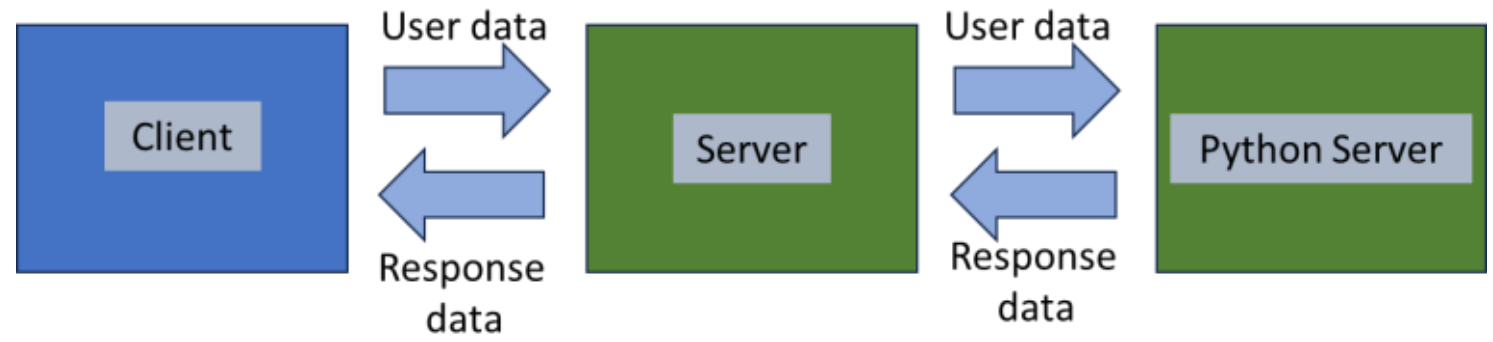

Figure 14: User request flow over the system

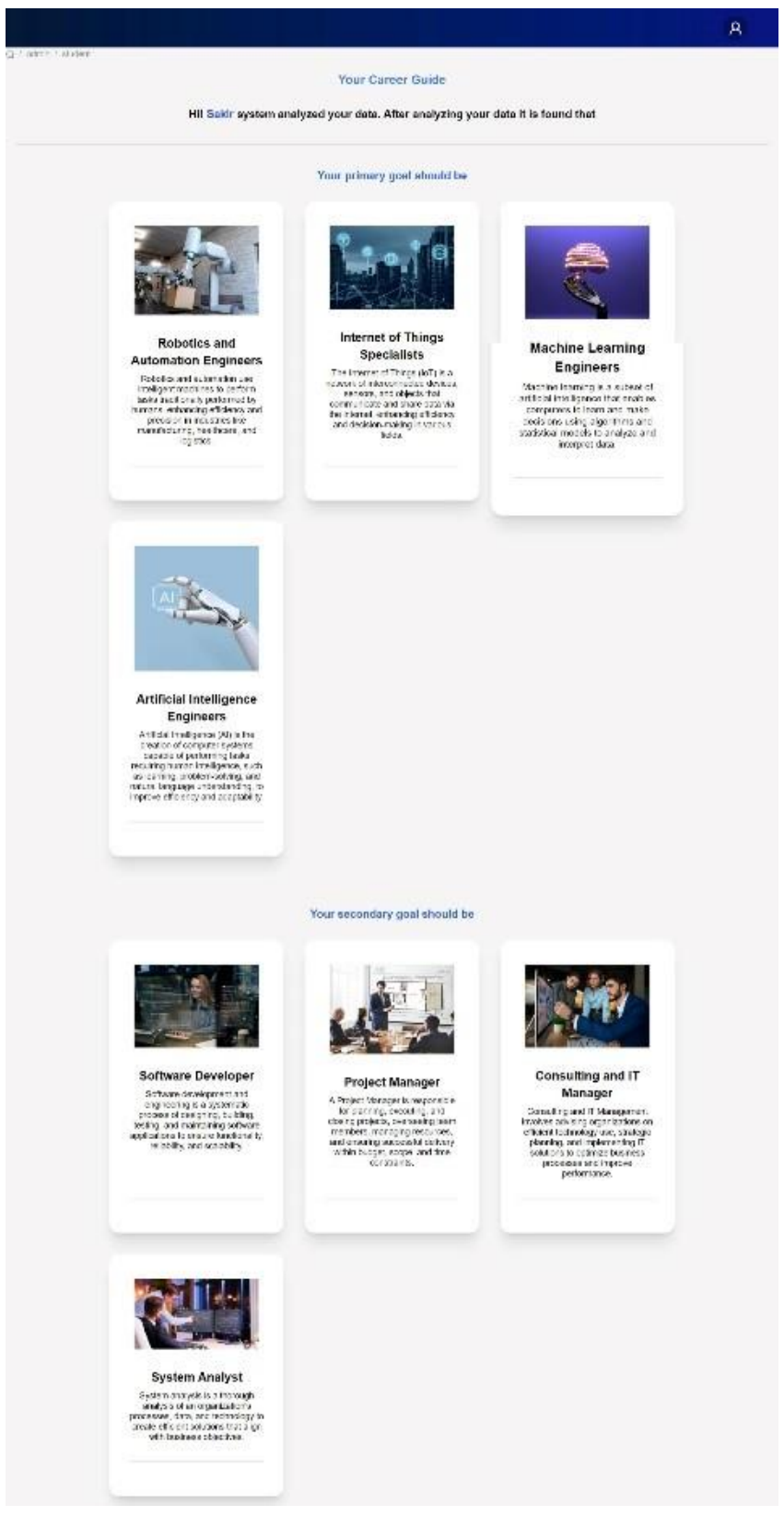

Figure 15: User interface presenting the predicted career fields generated by the system, adapted from (Faruque et al., 2025) with permission

# 6. Discussion

This project developed and evaluated a career guidance expert system for IT students, bridging the gap between academic performance, personal interests, and suitable career paths while offering web-based evaluations for personalized feedback. Our results show that the system provides accurate career recommendations, enhances student engagement, and helps align their skills with appropriate career opportunities.

## 6.1. Gap analysis and comparative discussion with previous studies

This study offers a unified approach that combines student assessment with career guidance—something previous research has not fully achieved. While many studies focus on assessments (Wang et al., 2004; Dahalan & Hussain, 2010; Coleman & Penuel, 2000; Antal & Koncz, 2011; Guzmán & Conejo, 2005) or career guidance (Supriyanto et al., 2019; Huang, 2022; Mehraj & Baba, 2019; Crişan et al., 2015; Myla et al., 2019), none integrate both into a single, complete system. Earlier works largely overlook skill-based evaluation for computer science students, instead emphasizing children's learning tests (Dahalan & Hussain, 2010), environmental awareness (Coleman & Penuel, 2000), or general self-assessment tools (Antal & Koncz, 2011). Our system addresses this by introducing targeted skill assessments tailored to IT students. Additionally, unlike previous career guidance models that rely mainly on surveys (Crişan et al., 2015; Myla et al., 2019), our approach considers fundamental skills, interests, and job-related experiences for more personalized recommendations. We also emphasize the role of teachers in guiding students—something rarely highlighted in past studies (Dahalan & Hussain, 2010; Coleman & Penuel, 2000; Antal & Koncz, 2011)—and prioritize system security through encrypted data storage and strong authentication, which previous research has largely ignored. Table 6 presents a comparative analysis of our study alongside the last research findings.

Table 6: Comparative Analysis with Previous Student Assessment and Career Guidance Technologies

| SL | Author Name | Proposed Method |
|---|---|---|
| 01 | Wang, Tzu-Hua, Kuo-Hua Wang, Wei-Lung Wang, Shih-Chieh Huang, and Sherry Y. Chen. et al. (Wang et al., 2004) | • Web-Based Test Analysis (WATA) system for teacher education<br>• The WATA system was designed and programmed based on test theory.<br>• Three steps are followed: Design, Programming of WATA system, and feedback from the user<br>• USSW survey was used to collect quantitative data. |
| 02 | Dahalan, Hamsiah Mohd, and Raja Maznah Raja Hussain et al. (Dahalan & Hussain, 2010) | • Develop e-ATLMS system and its functionality<br>• Taking users' feedback about the system's overall activities<br>• Highlighting the strengths and weaknesses of the e-ATLMS system |
| 03 | Coleman, Elaine B., and William R. Penuel. et al. (Coleman & Penuel, 2000) | • Develop web-based assessment to determine students' environmental |
| | | awareness and data analysis skills<br>• Data Collection from middle and high school students<br>• Based on students' responses environmental data is interpreted |
| 04 | Antal, Margit, and Szabolcs Koncz. et al. (Antal & Koncz, 2011) | • CBT system generates questions based on students' knowledge<br>• CAT system to analyze student behavior<br>• Computer-based test system<br>• Graphical representation of student knowledge |
| 05 | Guzmán, Eduardo, and Ricardo Conejo et al. (Guzmán & Conejo, 2005) | • SIETTE system that is used for administrative self-assessment CATS<br>• Item Response Theory (IRT) is a model that helps to select the next item and determine the test finish time |
| 06 | Supriyanto, G., Widiaty, I., Abdullah, A.G. and Yustiana, Y.R et al. (Supriyanto et al., 2019) | • Utilizes a systematic approach to analyze expert systems in educational guidance, evaluation, academic career guidance, and work guidance.<br>• Uses a spreadsheet program for analysis and coding of articles.<br>• Identifies relevant articles using search keywords like "expert systems, education, student guidance career, artificial intelligence."<br>• Extracts key findings and insights from selected articles.<br>• Uses a literature review spreadsheet model for structuring and summarizing information. |
| 07 | Huang, Li. et al. (Huang, 2022) | • The system combines algorithmic recommendation with ideological and political education<br>• Aggregate the preferences of students and employers<br>• Integrates student career competence |
| | | and guidance system<br>• Design a similarity-based random negative sampling module to address negative sample disbelief<br>• Includes two steps of similarity calculation between students and units<br>• Presents calculation method for inter-student and interunit similarity |
| 08 | Mehraj, Tehseen, and A. Mehraj Baba et al. (Mehraj & Baba, 2019) | • Examine AI schemes in career guidance and counseling<br>• Evaluates the scope of various technologies<br>• Discusses J48 decision tree algorithm and Case-Based Reasoning (CBR) techniques<br>• Acknowledge a system for career selection in education<br>• The system considers gender, IQ, grades, hobbies, and skills<br>• CBR algorithm has higher accuracy (80%) than J48 (50-60%).<br>• The system is limited to a specific geographical area and lacks a generic solution |
| 09 | Crişan, Claudia, Anişoara Pavelea, and Oana Ghimbuluţ. et al. (Crişan et al., 2015) | • Collect students' career information<br>• Surveying students' professional paths involves gathering insights into their career decisions, values, abilities, competencies, and future intentions.<br>• Participants were given a group briefing before completing the anonymous 30-minute paper-pencil questionnaire.<br>• Mixed methodology for a comprehensive understanding of students' career counseling needs |
| 10 | Myla, Saiteja, Surya Teja Marella, A. Swarnendra Goud, S. Hasane Ahammad, G. N. S. Kumar, and Syed Inthiyaz. et al. (Myla et al., 2019) | • Developed a rule-based and artificial intelligence system for career prediction<br>• Utilizes questionnaires to analyze |

students' mental abilities

- Conducts sample tests with completed higher secondary students
- Uses the expert system's output to allocate branches to students
- Represents data through graphs and tables
- Considers academic performance and provides branch recommendations

## 6.2. System sustainability

At present, students and faculty across several universities in Bangladesh are getting hands-on with the system. It's still in testing, so we're keeping a close eye on how it performs—things like speed, error rates, and how much it can handle at once. We're especially focused on keeping user data safe and paying attention to the feedback we get, since that's what shapes our next steps. To check how the system holds up under pressure, we ran tests using Apache JMeter and BlazeMeter. Table 7 lays out what happened when we pushed it with simulated loads. The system runs on AWS, too, which helps us keep things reliable and secure, building on the safety measures we talked about earlier in Section 4.2.3.

Table 7: System scalability measure based on total samples, throughput, latency range and error rate

| Sr. | Features | Sample No. | Throughput | Latency Range Over Request | Error rate |
|---|---|---|---|---|---|
| 01 | Chat | 4 | 1.9/sec | 1837–558 ms | 0.00% |
| 02 | Login | 11 | 3.2/sec | 658-4 ms | 0.00% |
| 03 | Skill Add | 2 | 1.2/sec | 2231 ms | 0.00% |
| 05 | Faculty Task Create | 2 | 1.5/sec | 1705 ms | 0.00% |
| 06 | The student provides solution | 7 | 2.8/sec | 797-7 ms | 0.00% |
| 07 | Faculty provides feedback on task | 5 | 1.1/sec | 1373-1158 ms | 0.00% |
| 08 | Task Solution Update | 4 | 1.7/sec | 734-883 ms | 0.00% |

As shown in Table 7, the system achieved a 0% error rate, though complex features like faculty feedback and task solution updates experienced higher latencies (734–1373 ms) due to intricate data interactions. Despite modest throughput during limited testing, the system remained stable and reliable, demonstrating readiness for further optimization and scaling.

## 6.3. Real world testing

In the WBSA and CGE systems, students and faculty are the main users, with students actively using core features and providing crucial feedback for usability assessment and future improvements. As shown in Figure 16 and reported in Faruque et al. (2025), 70% of students were satisfied with the skill assessment feature, while 90% reported a positive experience with interest- and skill-based activities, offering valuable insights into the system's effectiveness and guiding areas for refinement.

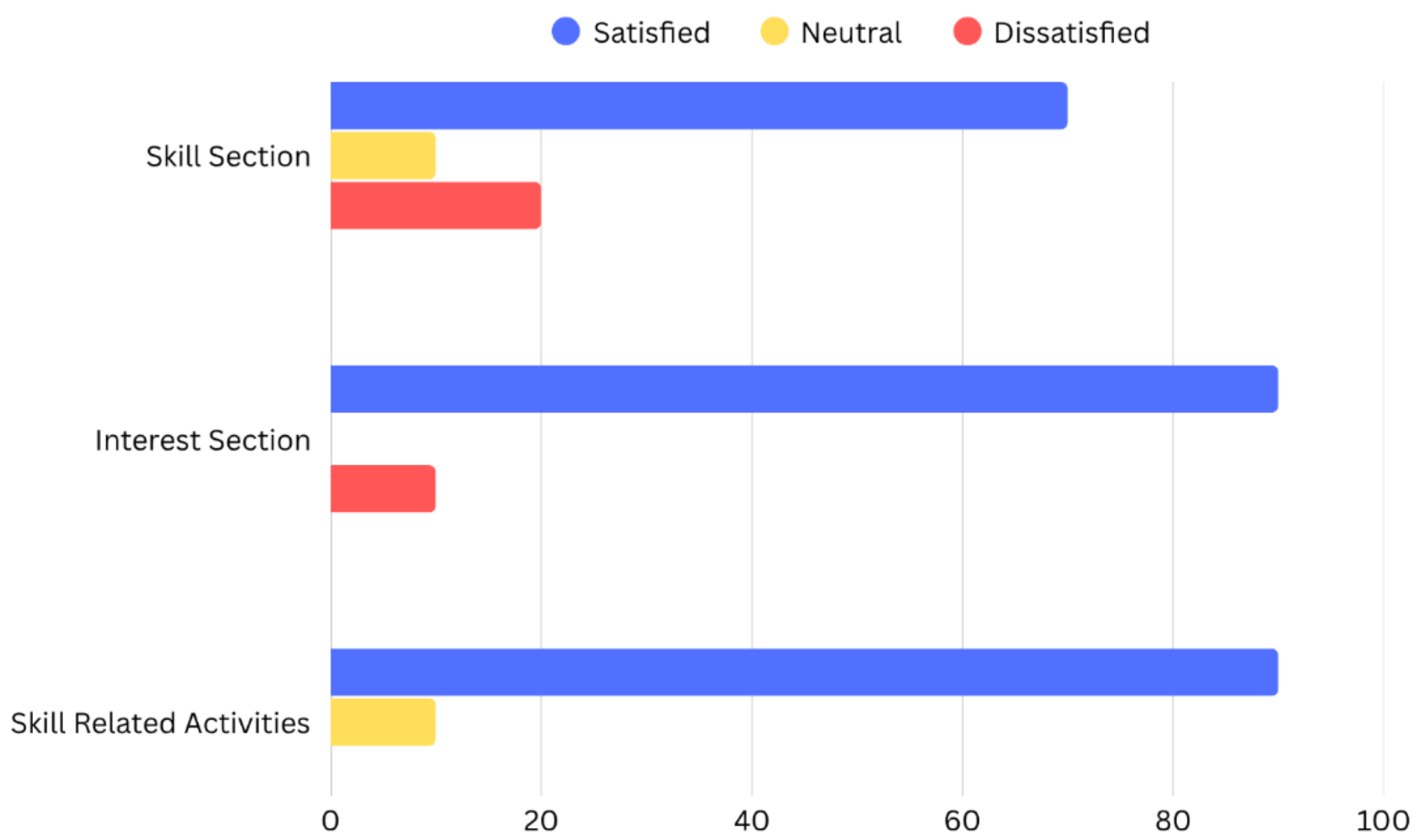

Figure 16: Students' feedback regarding the core functionalities of the system modified from (Faruque et al., 2025)

In our previous study (Faruque et al., 2025), the MLP model for career prediction proved highly accurate, correctly identifying suitable career paths in 90% of cases (Figure 17). Only 10% of predictions were inaccurate, demonstrating the model's practical reliability and value as a tool for student career guidance.

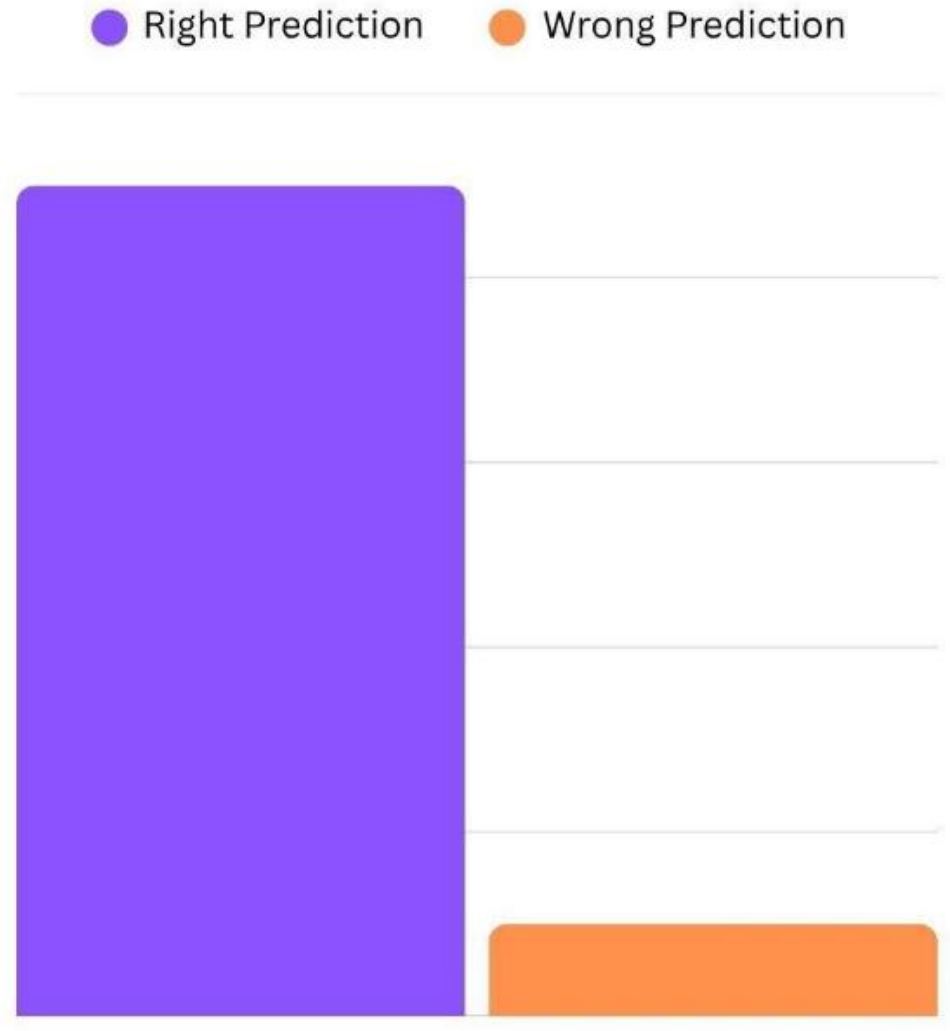

Figure 17: Graphical representation of the system's predicted career category statistics, modified from (Faruque et al., 2025)

Figure 18 summarizes faculty feedback on the dashboard, collected via a structured survey, covering system performance, functionality, UI/UX, feature needs, and potential improvements. Overall satisfaction was high for tasks, assignments, and interface experience, though some dissatisfaction arose with "Expertise Related Work," indicating room for minor enhancements while confirming the dashboard's effectiveness and usability.

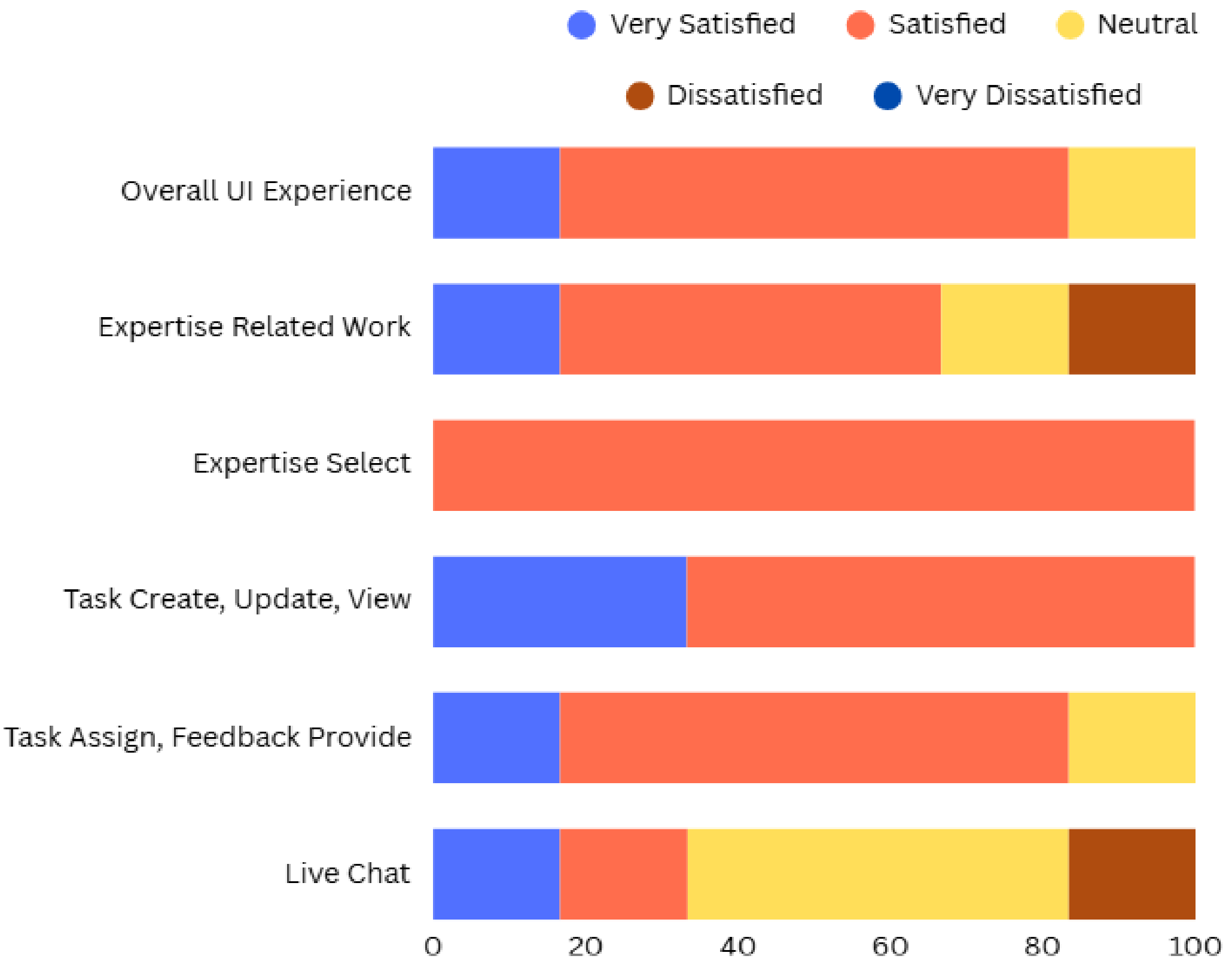

Figure 18: Faculty Feedback on Dashboard Features Based on System Usability Survey

# 7. Conclusion

This study shows that expert systems do more than just help students with their coursework—they actually guide them in making smarter choices about their careers. By bringing together the WBSA and CGE systems, the platform lets CS and SWE students check out career paths that really fit their strengths and interests, all while they build the skills they need. Faculty get helpful tools too: they can offer more tailored advice and keep an eye on how students are doing, which opens the door for more teamwork between students and instructors on research projects. Out of all the models we tried, the MLP came out on top for accuracy and now drives the career prediction feature. The platform also uses NLP to get a better grasp of what students care about and the skills that matter to them, making the advice it gives even more personal—for both students and faculty.

## 7.1. Limitations

The study was limited to IT students in CS and SWE, leaving its applicability to other disciplines untested. System performance was evaluated with a small user sample, which may not fully reflect behavior under high load, leaving potential latency or performance issues when scaled to larger user bases.

## 7.2. Future work

Future efforts aim to extend the system to other departments such as EEE, BBA, and English, offering personalized evaluation and career guidance across disciplines. We plan to improve the interface for ease of use, enhance scalability for more users, release a beta version for real-world feedback, and refine the prediction algorithm to provide more accurate and relevant job recommendations in line with evolving student needs and industry demands.

### Abbreviation:

Artificial Intelligence: AI

Data Science: DS

Development: DEV

Security: SEC

Software Development and Engineering: SDE

User Experience (UX) and User Interface (UI) Design: UI / UX

Career Prediction: CP

Multilayer Perceptron: MLP

Computer Science: CS

Software Engineering: SWE

Natural Language Processing: NLP

Neural Network: NN

Artificial Neural Networks: ANN

Deep Learning: DL

Machine Learning: ML

Web-Based Student Assessment – WBSA

Career Guidance Expert – CGE

## Reference

Aadmin. (2024, September 4). What is Naviance? Everything you need to know. Yocket. https://yocket.com/us/blog/what-is-naviance

Antal, M., & Koncz, S. (2011). Student modeling for a web-based self-assessment system. Expert Systems with Applications, 38(6), 6492–6497. https://doi.org/10.1016/j.eswa.2010.11.118

Bandura, A., Barbaranelli, C., Caprara, G. V., & Pastorelli, C. (2001). Self-efficacy beliefs as aspirations and career trajectories. Child Development, 72(1), 187–206. https://doi.org/10.1111/1467-8624.00273

Coleman, E. B., & Penuel, W. R. (2000). Web-based student assessment for program evaluation. Journal of Science Education and Technology, 9, 327–342. https://doi.org/10.1023/A:1009481505429

Crişan, C., Pavelea, A., & Ghimbuluţ, O. (2015). A need assessment on students' career guidance. *Procedia-Social and Behavioral Sciences, 180*, 1022–1029. https://doi.org/10.1016/j.sbspro.2015.02.191

Dahalan, H. M., & Hussain, R. M. R. (2010). Development of web-based assessment in teaching and learning management system (e-ATLMS). *Procedia-Social and Behavioral Sciences, 9*, 244–248. https://doi.org/10.1016/j.sbspro.2010.12.144

Doguslu, S. S. (2024, November 27). Qooper: Features, price, reviews & rating - eLearning Industry. eLearning Industry. https://elearningindustry.com/directory/elearning-software/qooper

Faruque, S.H., Khushbu, S.A. & Akter, S. Decision support system to reveal future career over students' survey using explainable AI. Educ Inf Technol 30, 14471–14509 (2025). https://doi.org/10.1007/s10639-025-13361-7

Guzmán, E., & Conejo, R. (2005). Self-assessment in a feasible, adaptive web-based testing system. IEEE Transactions on Education, 48(4), 688–695. https://doi.org/10.1109/TE.2005.852591

Hirschi, A. (2018). The Fourth Industrial Revolution: Issues and implications for career research and practice.

Career Development Quarterly, 66(2), 192–204. https://doi.org/10.1002/cdq.12142
Huang, L. (2022). The establishment of college student employment guidance system integrating artificial intelligence and civic education. Mathematical Problems in Engineering, 2022. https://doi.org/10.1155/2022/3934381
Kazi, A. S., & Akhlaq, A. (2017). Factors affecting students' career choice. Journal of Research and Reflections in Education, 2(2), 187–196.
Mehraj, T., & Baba, A. M. (2019). Scrutinizing artificial intelligence based career guidance and counselling systems: An appraisal. International Journal of Interdisciplinary Research and Innovations, 7(1), 402–411.
Myla, S., Marella, S. T., Goud, A. S., Ahammad, S. H., Kumar, G. N. S., & Inthiyaz, S. (2019). Design decision taking system for student career selection for accurate academic system. International Journal of Scientific and Technology Research, 8(9), 2199–2206.
Nyamwange, J. (2016). Influence of student's interest on career choice among first year university students in public and private universities in Kisii County, Kenya. Journal of Education and Practice, 7(4), 96–102.
Our mission | Year Up United. (2025). Year Up. https://www.yearup.org/about/our-mission
Rangnekar, R. H., Suratwala, K. P., Krishna, S., & Dhage, S. (2018, August). Career prediction model using data mining and linear classification. In 2018 Fourth International Conference on Computing Communication Control and Automation (ICCUBEA) (pp. 1–6). IEEE. https://doi.org/10.1109/ICCUBEA.2018.8697392
Razak, T. R., Hashim, M. A., Noor, N. M., Halim, I. H. A., & Shamsul, N. F. F. (2014). Career path recommendation system for UiTM Perlis students using fuzzy logic. In 2014 5th International Conference on Intelligent and Advanced Systems (ICIAS) (pp. 1–5). IEEE. https://doi.org/10.1109/ICIAS.2014.6869553
Reddy, M. M. (2021). Career prediction system. International Journal of Scientific Research in Science and Technology, 8(4), 54–58.
Shahiri, A. M., & Husain, W. (2015). A review on predicting student's performance using data mining techniques. Procedia Computer Science, 72, 414–422. https://doi.org/10.1016/j.procs.2015.12.157
Supriyanto, G., Widiaty, I., Abdullah, A. G., & Yustiana, Y. R. (2019, December). Application expert system career guidance for students. In Journal of Physics: Conference Series (Vol. 1402, No. 6, p. 066031). IOP Publishing. https://doi.org/10.1088/1742-6596/1402/6/066031
University of Technology Sydney. (2025). Professional mentoring platform. https://www.uts.edu.au/for-students/current-students/support/opportunities/careers/featured-tools-and-resources/professional-mentoring-platform
Wang, T. H., Wang, K. H., Wang, W. L., Huang, S. C., & Chen, S. Y. (2004). Web-based Assessment and Test Analyses (WATA) system: Development and evaluation. Journal of Computer Assisted Learning, 20(1), 59–71. https://doi.org/10.1111/j.1365-2729.2004.00070.x
Wikipedia contributors. (2025, April 25). Naviance. Wikipedia. https://en.wikipedia.org/wiki/Naviance